\documentclass[10pt]{article}   

\usepackage{floatrow} \newfloatcommand{capbtabbox}{table}[][\FBwidth]

\usepackage[utf8]{inputenc}

\usepackage{fancyhdr}
\usepackage{xcolor,colortbl}
\usepackage{listings}
\usepackage{soul}
\usepackage{url}
\usepackage[hidelinks]{hyperref}
\definecolor{mediumpersianblue}{rgb}{0.0, 0.4, 0.65}
\definecolor{prussianblue}{rgb}{0.0, 0.27, 0.46}
\definecolor{darkpersianblue}{rgb}{0.04, 0.11, 0.23}
\hypersetup{pdftex,colorlinks=true,allcolors=prussianblue,pdfpagemode=UseOutlines}

\usepackage{hypcap}
\usepackage[utf8]{inputenc}
\usepackage[small]{caption}
\usepackage{graphicx}
\usepackage{adjustbox}
\usepackage{tabularx}
\usepackage{amsmath}
\usepackage{amsthm}
\usepackage{amsfonts}
\usepackage{amssymb}
\usepackage{mathtools}
\usepackage{commath}
\usepackage{usebib}
\usepackage{bm}
\usepackage{makecell}
\usepackage{chngcntr}  \usepackage{booktabs}
\usepackage{etoolbox}
\usepackage{multirow}
\usepackage{xspace}
\usepackage{algorithm}
\let\classAND\AND
\let\AND\relax
\usepackage{algpseudocode}
\let\AND\classAND
\AtBeginEnvironment{algorithmic}{\let\AND\algoAND}

\usepackage[english]{selnolig}

\usepackage{scrlayer}
\DeclareNewLayer[
  foreground,
  topmargin,
  addheight=\dimexpr\headheight+\headsep\relax,
  contents=\layercontentsmeasure
]{measurelayer}
\DeclareNewPageStyleByLayers{measurestyle}{measurelayer} 

\usepackage{amsmath,amsfonts,bm}

\def\eqref#1{equation~\ref{#1}}

\def\floor#1{\lfloor #1 \rfloor}
\def\1{\bm{1}}

\DeclareMathAlphabet{\mathsfit}{\encodingdefault}{\sfdefault}{m}{sl}
\SetMathAlphabet{\mathsfit}{bold}{\encodingdefault}{\sfdefault}{bx}{n}

\def\gH{{\mathcal{H}}}

\DeclareMathOperator*{\E}{\mathbb{E}}

 \usepackage{hyphenat}
\let\origaddcontentsline\addcontentsline
\usepackage[accepted]{tmlr}  \let\addcontentsline\origaddcontentsline

\graphicspath{{../figures/}{./figures}{.}{../}{..}}
\usepackage[labelformat=simple]{subcaption}

\title{Action Noise in Off-Policy Deep Reinforcement Learning: Impact on Exploration and Performance}

\author{\name Jakob Hollenstein \email jakob.hollenstein@uibk.ac.at\\
  \addr Department of Computer Science, University of Innsbruck
  \AND
  \name Sayantan Auddy \email sayantan.auddy@uibk.ac.at\\
  \addr Department of Computer Science, University of Innsbruck
  \AND
  \name Matteo Saveriano \email matteo.saveriano@unitn.it\\
  \addr Department of Industrial Engineering, University of Trento
  \AND
  \name Erwan Renaudo \email erwan.renaudo@uibk.ac.at\\
  \addr Department of Computer Science, University of Innsbruck
  \AND
  \name Justus Piater \email justus.piater@uibk.ac.at\\
  \addr Department of Computer Science, University of Innsbruck
}

\newcommand{\myparagraph}[1]{\par\textbf{#1}\space}
\makeatletter
\let\oldmyparagraph\myparagraph
\renewcommand\myparagraph[1]{\oldmyparagraph{#1}\@ifnextchar\par{\@gobble}{}}
\makeatother

\date{Feb 2022}

\usepackage{natbib}

\renewcommand*{\cite}{\citep}
\bibinput{pps-exploration-metric} 
\usepackage[inline]{enumitem}
\usepackage{graphicx}
\usepackage{amsmath}
\usepackage{amssymb}
\usepackage{url}
\usepackage{soul}
\usepackage{verbatim}
\usepackage{environ,censor,calc}
\NewEnviron{mysout}
 {\soutrefunexp{\BODY}}
\makeatletter
\renewcommand\@cenword[1]{\setlength{\nextcharwidth}{\widthof{#1}}\censorrule{\nextcharwidth}\kern -\nextcharwidth #1}
\makeatother
\newcommand\soutref[1]{\censorruledepth=.55ex\xblackout{#1}}
\newcommand\soutrefunexp[1]{\expandafter\soutref\expandafter{#1}}

\usepackage{xcolor}
\usepackage{colortbl}

\definecolor{orchid}{rgb}{0.85, 0.44, 0.84}
\definecolor{palecopper}{rgb}{0.85, 0.54, 0.4}
\definecolor{sapgreen}{rgb}{0.31, 0.49, 0.16}
\definecolor{amber(sae/ece)}{rgb}{1.0, 0.49, 0.0}
\definecolor{mediumpersianblue}{rgb}{0.0, 0.4, 0.65}

\usepackage{xparse}
\usepackage{xargs}

\usepackage[colorinlistoftodos,prependcaption,textsize=tiny]{todonotes}

\newcommandx{\jhnotedone}[2][1=]{\todo[linecolor=green!60!black,backgroundcolor=green!60!black!25,bordercolor=green!60!black,#1]{\st{#2}}}

\newcommandx{\jhtododone}[2][1=]{\todo[linecolor=amber(sae/ece),backgroundcolor=amber(sae/ece)!25,bordercolor=amber(sae/ece),#1]{\st{#2}}}

\newcommandx{\ercommentdone}[2][1=]{\todo[linecolor=palecopper,backgroundcolor=palecopper!25,bordercolor=palecopper,#1]{\st{#2}}}

\newcommandx{\mscommentdone}[2][1=]{\todo[linecolor=orchid,backgroundcolor=orchid!25,bordercolor=orchid,#1]{\st{#2}}}

\newcommandx{\sacommentdone}[2][1=]{\todo[linecolor=orchid,backgroundcolor=orchid!25,bordercolor=orchid,#1]{\st{#2}}}

\newcommandx{\revcommentdone}[2][1=]{\todo[linecolor=amber(sae/ece),backgroundcolor=amber(sae/ece)!25,bordercolor=orchid,#1]{\st{#2}}}

\renewcommand{\refeq}[1]{(\ref{#1})}
\newcommand{\refalg}[1]{Algorithm~\ref{#1}}
\newcommand{\reftbl}[1]{Table~\ref{#1}}
\newcommand{\refsec}[1]{Section~\ref{#1}}
\newcommand{\reffig}[1]{Figure~\ref{#1}}
\newcommand{\xurel}{\operatorname{X}_{\mathcal{U}\text{rel}}}
\newcommand{\clip}{\operatorname*{clip}}
\newcommand{\DKL}{D_{\textrm{KL}}}
\newcommand{\xbin}{X_\text{bin}}
\newcommand{\xbbm}{X_\text{BBM}}
\newcommand{\xnn}{X_\text{NN}}

\newcommand{\citeintext}[1]{\citet{#1}}  \newcommand{\libname}[1]{\texttt{#1}}

\def\month{11}  \def\year{2022} \def\openreview{\url{https://openreview.net/forum?id=NljBlZ6hmG}} \makeatletter
\begin{document}

\makeatletter
\let\usetitle\@title
\makeatother
\maketitle
\makeatletter
\edef\textFontName{\fontname\csname
  \f@encoding/\f@family/\f@series/\f@shape/\f@size\endcsname}
\edef\mathFontName{\fontname\textfont0}
\edef\mathLetterFontName{\fontname\textfont1}
\makeatother
\definecolor{colGaussBetter}{rgb}{1.,0.8,0.55} \definecolor{colOUBetter}{rgb}{0.8,1,0.8} \definecolor{ColSac}{rgb}{0.8,0.5,0.3} \definecolor{ColDetsac}{rgb}{0.3,.5,0.0} \newcommand{\syntheticDatadim}{25}
\newcommand{\vP}{$P$\xspace}
\newcommand{\vR}{$R$\xspace}
\newcommand{\vX}{$X$\xspace}
\newcommand{\vE}{$E$\xspace}
\newcommand{\vvarP}{$\textrm{var}({P})$\xspace}
\providecommand{\change}[1]{#1}
\providecommand{\delchange}[1]{\ignorespaces}

\begin{abstract}
  Many Deep Reinforcement Learning (D-RL) algorithms rely on simple
  forms of exploration such as the additive action noise often used
  in continuous control domains.  Typically, the scaling factor of
  this action noise is chosen as a hyper-parameter and is kept constant
  during training. In this paper, we focus on action noise in
  off-policy deep reinforcement learning for continuous control. We
  analyze how the learned policy is impacted by the noise type, noise scale,
  and impact scaling factor reduction schedule.
We consider the two
  most prominent types of action noise, Gaussian and
  Ornstein-Uhlenbeck noise, and perform a vast experimental campaign
  by systematically varying the noise type and scale parameter, and by
  measuring variables of interest like the expected return of the
  policy and the state-space coverage during exploration. For the
  latter, we propose a novel state-space coverage measure $\xurel$
  that is more robust to \change{estimation artifacts caused by points
  close to the state-space boundary} \delchange{boundary artifacts} than
  previously-proposed measures.
Larger noise scales generally increase state-space
  coverage. However, we found that increasing the space coverage using
  a larger noise scale is often not beneficial. On the contrary,
  reducing the noise scale over the training process reduces the
  variance and generally improves the learning performance.  We
  conclude that the best noise type and scale are environment
  dependent, and based on our observations derive heuristic rules for guiding
  the choice of the action noise as a starting point for further
  optimization. \url{https://github.com/jkbjh/code-action_noise_in_off-policy_d-rl}
\end{abstract}

\section{Introduction }
In (deep) reinforcement learning an agent aims to learn a policy to
act optimally based on data it
collects by interacting with the environment. In order to learn a well
performing policy, data (i.e.\ state-action-reward sequences) of
sufficiently good behavior need to be collected. A simple and very
common method to discover better data is to induce variation in the
data collection by adding noise to the action selection
process. Through this variation, the agent will try a wide range of action
sequences and eventually discover useful information.

\myparagraph{Action Noise} In off-policy reinforcement learning algorithms applied to continuous
control domains, a go-to approach is to add a randomly-sampled
\emph{action noise} to the action chosen by the policy.
Typically the action noise is sampled from a Gaussian distribution or
an Ornstein-Uhlenbeck process, either because algorithms are proposed
using these noise types
\cite{fujimotoAddressingFunctionApproximation2018,lillicrapContinuousControlDeep2016},
or because these two types are provided by reinforcement learning
implementations
\cite{liangRLlibAbstractionsDistributed2018,stable-baselines3,fujitaChainerRLDeepReinforcement2021,senoD3rlpyOfflineDeep2021}. 
While adding action noise is simple, widely used, and surprisingly
effective, the impact of action noise type or scale does not
feature very prominently in the reinforcement learning
literature. However, the action noise can have a huge impact on the
learning performance as the following example shows.
\myparagraph{A motivating example\label{sec:motivating_example}:}

Consider the case of the \emph{Mountain-Car}
\change{\cite{brockmanOpenAIGym2016,
    mooreEfficientMemorybasedLearning1990}} environment.
In this task, a car starts in a valley between mountains on the left
and right and does not have sufficient power to simply drive up the
mountain. It needs repetitive swings to increase its potential and
kinetic energy to finally make it up to the top of the mountain on the
right side. The actions apply a force to the car and incur a cost that
is quadratic to the amount of force, while reaching the goal yields a
final reward of 100. This parameterization implies a local optimum:
not performing any action and achieving a return of zero.

\begin{figure}
  \floatsetup{heightadjust=object}  A motivating example: Mountain-Car
  \begin{floatrow}

\capbtabbox[0.4\textwidth]{\begin{tabular}{lrr}
      \toprule
      \makecell{noise \\Type}         &
      Gaussian           &
                           \makecell{Ornstein-\\Uhlenbeck} \\
      \midrule
 Scale & 0.6 & 0.5 \\
 Return  & -30.2 & -30.4 \\
 +-    & 0.1 & 1.3 \\
  \bottomrule
    \end{tabular}
  \vspace{.23em}
}{\caption{Untrained random policies, Gaussian
    ($\sigma=0.6$) and Ornstein-Uhlenbeck ($\sigma=0.5$) achieve
    similar returns and appear interchangeable. }
  \label{tbl:mountaincar_motivation}
}
\ffigbox[0.59\textwidth]{\includegraphics[width=\linewidth]{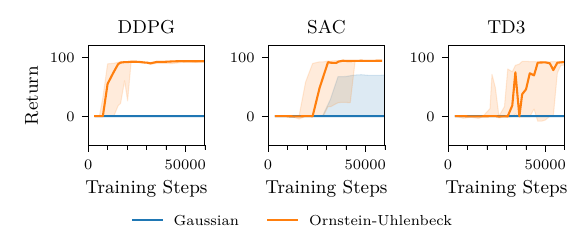}

}{\caption{Training with the action noises
    (\reftbl{tbl:mountaincar_motivation}) shows the impact of
      noise type; Ornstein-Uhlenbeck solves the task, but Gaussian
      does not. Other algorithm parameters are taken from the tuned
      parameters found by \citeintext{rl-zoo3}.  The lines indicate the
      medians, the shaded areas the quartiles of ten independent
      runs.}
  \label{fig:mountaincar_motivation_learning}
}
\end{floatrow}

\end{figure}

Driving the environment with purely random policies based on the two
noise types (Gaussian, $\sigma=0.6$, Ornstein-Uhlenbeck $\sigma=0.5$,
see \reftbl{tbl:mountaincar_motivation}), yields similar returns.
However, when we apply the algorithms DDPG, TD3 and SAC
\cite{lillicrapContinuousControlDeep2016,
  fujimotoAddressingFunctionApproximation2018,
  haarnojaSoftActorCriticAlgorithms2019} to this task, the resulting
learning curves (\reffig{fig:mountaincar_motivation_learning}) very
clearly depict the huge impact the noise configuration has. While
returns of the purely random noise-only policies were similar, we
achieve substantially different learning results. Learning either
fails (Gaussian) or leads to success (Ornstein-Uhlenbeck). This shows
the huge importance of the action noise configuration. See
\refsec{sec:motivating_example_details} for further details.

\change{\myparagraph{Exploration Schedule}}A very common strategy in Q-learning algorithms applied to discrete
control is to select a random action with a certain probability
$\varepsilon$. In this \emph{epsilon-greedy} strategy, the probability
$\varepsilon$ is often chosen higher in the beginning of the training
process and reduced to a smaller value over course of the training.
Although very common in Q-learning, a comparable strategy has not
received a lot of attention for action noise in continuous control.
The \change{descriptions} of the most prominent algorithms using action noise, namely DDPG
\cite{lillicrapContinuousControlDeep2016} and TD3
\cite{fujimotoAddressingFunctionApproximation2018}, do not mention
changing the noise over the training process. Another prominent
algorithm, SAC \cite{haarnojaSoftActorCriticAlgorithms2019}, adapts
the noise to an entropy target. The entropy target, however, is kept
constant over the training process. In many cases the optimal policy
would be deterministic, but the agent has to behave with similar
average action-entropy no matter whether the optimal policy has been
found or not. 

\change{An} indication that reducing the randomness over the training
process has received little attention is that only very few
reinforcement learning implementations, e.g.,
RLlib~\cite{liangRLlibAbstractionsDistributed2018}, implement reducing
the impact of action noise over the training progress. Some libraries,
like \texttt{coach}~\cite{caspiReinforcementLearningCoach2017}, only
implement a form of continuous epsilon greedy: sampling the action
noise from a uniform distribution with probability $\varepsilon$. The
majority of available implementations, including
\libname{stable-baselines}~\cite{stable-baselines3},
\libname{PFRL}~\cite{fujitaChainerRLDeepReinforcement2021},
\libname{acme}~\cite{hoffmanAcmeResearchFramework2020}, and
\libname{d3rlpy}~\cite{senoD3rlpyOfflineDeep2021}, do not implement
any strategies to reduce the impact of action noise over the training
progress.

\change{Exploration schedules for action noise are also not mentioned
  in several recent surveys
  \citep{yangExplorationDeepReinforcement2022,
    ladoszExplorationDeepReinforcement2022,
    aminSurveyExplorationMethods2021}
  }

\myparagraph{Contributions}

In this paper we analyze the impact of Gaussian and Ornstein-Uhlenbeck
noise on the learning process of DDPG, TD3, SAC and a deterministic
SAC variant. Evaluation is performed on multiple popular environments
(\reftbl{tbl:envimgs}): Mountain-Car~\cite{brockmanOpenAIGym2016}
environment from the OpenAI Gym, Inverted-Pendulum-Swingup, Reacher,
Hopper, Walker2D and Half-Cheetah environments implemented using
PyBullet \cite{coumansPyBulletPythonModule2016,benelot2018}.
\begin{itemize}
\item We investigate the relation between exploratory state-space
  coverage \vX, returns collected by the exploratory policy \vR and
  learned policy performance \vP.

\item We propose to assess the state-space coverage using our novel
  measure $\xurel$ that is more robust to approximation artifacts on
  bounded spaces compared to previously proposed measures.

\item We perform a vast experimental study and investigate the
  question whether one of the two noise types is generally preferable
  \emph{(Q1)}, whether a specific scale should be used \emph{(Q2)},
  whether there is any benefit to reducing the scale over the training
  progress (linearly, logistically) compared to keeping it constant
  \emph{(Q3)}, and which of the parameters noise type, noise scale and
  scheduler is most important \emph{(Q4)}.

\item We provide a set of heuristics derived from our results to guide
  the selection of initial action noise configurations.

\end{itemize}

\myparagraph{Findings} We found that the noise configuration, noise
type and noise scale, have an important impact and can be necessary
for learning (e.g.\ Mountain-Car) or can break learning (e.g.\
Hopper). Larger noise scales tend to increase state-space coverage,
but for the majority of our investigated environments increasing the
state-space coverage is not beneficial\change{: increased state-space
  coverage was associated with a reduction in performance. This
  indicates that in these environments, local exploration, which is
  associated with smaller state-space coverage, tends to be favorable.}
\emph{We recommend to select and tune action noise based on the reward
  and dynamics structure on a per-environment basis.}

We found that across noise configurations, decaying the impact of
action noise tends to work better than keeping the impact constant, in
both reducing the variance across seeds and improving the learned
policy performance and can thus make the algorithms more robust to the
action noise hyper-parameters scale and type. \emph{We recommend to
  reduce the action noise scaling factor over the training time.}

We found that for all environments investigated in this study
noise scale $\sigma$ is the most important parameter, and some
environments (e.g.\ Mountain-Car) benefit from larger noise scales,
while other environments require very small scales (e.g.\
Walker2D). \emph{We recommend to assess an environment's action noise
  scale preference first}.

\section{Related Work}

By combining Deep Learning with Reinforcement Learning in their DQN
method, \citeintext{mnihHumanlevelControlDeep2015} achieved substantial
improvements on the Atari Games RL benchmarks
\cite{bellemareArcadeLearningEnvironment2013} and sparked lasting
interest in \emph{Deep Reinforcement learning} (D-RL).
\myparagraph{Robotic environments:}

In robotics, the interest in Deep Reinforcement Learning has also been
rising and common benchmarks are provided by OpenAI
Gym~\cite{brockmanOpenAIGym2016}, which includes control classics such
as the Mountain-Car
environment~\cite{mooreEfficientMemorybasedLearning1990} as well as
more complicated (robotics) tasks based on the Mujoco
simulator~\cite{todorovMuJoCoPhysicsEngine2012}. Another common
benchmark is the DM Control Suite~\cite{tassaDeepMindControlSuite2018}, also based on Mujoco. While
Mujoco has seen widespread adoption it was, until recently, not freely
available. A second popular simulation engine, that has been freely
available, is the Bullet simulation
engine~\cite{coumansPyBulletPythonModule2016} and very similar
benchmark environments are also available for the Bullet 
engine~\cite{coumansPyBulletPythonModule2016,benelot2018}.

\myparagraph{Continuous Control:} While the Atari games feature large
and (approximately) continuous observation spaces, their action spaces
are discrete and relatively small, making Q-learning a viable
option. In contrast, typical robotics tasks require \emph{continuous
  action spaces}, implying uncountably many different actions.

A tabular Q-learning approach or a discrete Q-function output for each
action are therefore not possible and maximizing the action over a
learned function approximator for $Q(s,a)$ is computationally
expensive (although not impossible as
\citeintext{kalashnikovQTOptScalableDeep2018} have shown). Therefore,
in continuous action spaces, \emph{policy search} is employed, to
directly optimize a function approximator \emph{policy}, mapping from
state to best performing
action~\cite{williamsSimpleStatisticalGradientfollowing1992}. To still
reap the benefits of reduced sample complexity of TD-methods, policy
search is often combined with learning a value function, a
\emph{critic}, leading to an \emph{actor-critic} approach
\cite{suttonPolicyGradientMethods1999}.

\myparagraph{On- and Off-policy:}

Current state of the art D-RL algorithms consist of \emph{on-policy}
methods, such as TRPO~\cite{schulmanTrustRegionPolicy2015} or
PPO~\cite{schulmanProximalPolicyOptimization2017}, and \emph{off-policy}
methods, such as DDPG~\cite{lillicrapContinuousControlDeep2016},
TD3~\cite{fujimotoAddressingFunctionApproximation2018} and
SAC~\cite{haarnojaSoftActorCriticAlgorithms2019}. While the on-policy
methods optimize the next iteration of the policy with respect to the data
collected by the current iteration, off-policy methods are, apart from
stability issues and requirements on the samples, able to improve
policy performance based on data collected by \emph{any arbitrary}
policy and thus can also re-use older samples.

To improve the policy, variation (\emph{exploration}) in the collected
data is necessary. The most common form of exploration is based on
randomness: in on-policy methods this comes from a \emph{stochastic
  policy} (TRPO, PPO), while in the off-policy case it is possible to
use a stochastic policy (SAC) or, to use a \emph{deterministic policy}
\cite{silverDeterministicPolicyGradient2014} with added \emph{action
  noise} (DDPG, TD3). Since off-policy algorithms can learn from data
collected by other policies, it is also possible to combine stochastic
policies (e.g. SAC) with action noise.

\myparagraph{State-Space Coverage:}

Often, the reward is associated with reaching certain areas in the
state-space. Thus, in many cases, \emph{exploration} is related to
\emph{state-space coverage}. An intuitive method to calculate state
space coverage is based on binning the state-space and counting the
percentage of non-empty bins. Since this requires exponentially many
points as the dimensionality increases, other measures are necessary.
\citeintext{zhanTakingScenicRoute2019} propose to measure state
coverage by drawing a bounding box around the collected data and
measuring the means of the side-lengths, or by measuring the sum of the
eigenvalues of the estimated covariance matrix of the collected
data. However, so far, there is no common and widely adopted approach.

\myparagraph{Methods of Exploration:}
The architecture for the stochastic policy in
SAC~\cite{haarnojaSoftActorCriticAlgorithms2019} consists of a neural
network parameterizing a Gaussian distribution, which is used to
sample actions and estimate action-likelihoods.
A similar stochastic policy architecture is also used in
TRPO~\cite{schulmanTrustRegionPolicy2015} and
PPO~\cite{schulmanProximalPolicyOptimization2017}. While this is the
most commonly used type of distribution, more complicated
parameterized stochastic policy distributions based on normalizing
flows have been
proposed~\cite{mazoureLeveragingExplorationOffpolicy2020,wardImprovingExplorationSoftActorCritic2019}.
In \change{the} case of action noise, the noise processes are not
limited to uncorrelated Gaussian (e.g. TD3) and temporally correlated
Ornstein-Uhlenbeck noise (e.g. DDPG): a whole family of action noise
types is available under the name of colored noise, which has been
successfully used to improve the
Cross-Entropy-Method~\cite{pinneriSampleefficientCrossEntropyMethod2020}.
A quite different type of random exploration are the parameter space
exploration methods \cite{maniaSimpleRandomSearch2018,
  plappertParameterSpaceNoise2018}, where noise is not applied to the
resulting action, but instead, the parameters of the policy are
varied. As a somewhat intermediate method, state dependent exploration
\cite{raffinSmoothExplorationRobotic2021} has been proposed, where
action noise is deterministically generated by a function based on
the state. Here, the function parameters are changed randomly for each
episode, leading to different deterministic ``action noise'' for each episode.
Presumably among the most intricate methods to generate exploration
are the methods that train a policy to achieve exploratory behavior
by rewarding exploratory actions \cite{burdaExplorationRandomNetwork2019,
  tangExplorationStudyCountBased2017, muttiPolicyGradientMethod2020,
  hongDiversitydrivenExplorationStrategy2018,
  pongSkewFitStateCoveringSelfSupervised2020}.
\change{Another alternative can be a two-step approach, where in the
  first stage intrinsically-motivated exploration is used to populate
  the replay buffer, and in the second stage the information in the
  buffer is exploited to learn a policy
  \cite{colasGEPPGDecouplingExploration2018}.}

It is however, not clear yet, which exploration method is most
beneficial, and when a more complicated method is actually worth the
additional computational cost and complexity. In this work we aim to
reduce this gap, by investigating the most widely used baseline method
in more detail: exploration by action noise.

\change{
\paragraph{Studies of Random Exploration}
Exploration in Deep Reinforcement Learning is also the subject of
multiple surveys. However, the topic of action noise is only covered very
sparsely.
\citeintext{yangExplorationDeepReinforcement2022} only brief\-ly mention
action noise as being used in DDPG
\citep{lillicrapContinuousControlDeep2016} and TD3
\citep{fujimotoAddressingFunctionApproximation2018} but do not provide
further discussion.  A section on randomness-based methods that focuses
mostly on discrete action spaces, or for continuous action spaces
on parameter noise, is provided by
\citeintext{ladoszExplorationDeepReinforcement2022}.
\citeintext{aminSurveyExplorationMethods2021} provide a section on
randomized action selection in policy search, nicely divided
into action-space and parameter-space methods, and discuss temporally
correlated or uncorrelated perturbations. However, they also do not
point to any empirical study specifically comparing the effects of
random exploration.

Generally, however, it appears that most work focuses on proposing
modifications of the action noise, rather than investigating the
effects of the baseline parameters. For example, for stochastic
policies, \citeintext{raoHowMakeDeep2020} show that the weight initialisation
procedure can lead to different initial action distributions of the
stochastic policies. \citeintext{chouImprovingStochasticPolicy2017} propose
stochastic policies based on the
$\beta$-distribution. \citeintext{nobakhtActionSpaceNoise2022} use gathered
experience to tune the action noise model.

In contrast to these works, in our previous work
  \citep{hollensteinHowDoesType2021}, we investigated action noise as the immediate means to control the environments, i.e.\ adding
  action noise to a constant-zero policy. Not surprisingly, we found that there are dependencies between environment dynamics, reward structure and action
  noise. However, this study did not investigate the influence of action noise on
  learning progress or results. In this work, we investigate the impact of action noise
  in the context of learning.
}
\section{Methods}

In this section, we describe the action noise types, the schedulers to
reduce the scaling factor of the action noise over time and the
evaluation process in more detail.  We brief\-ly list the analyzed
benchmark environments and their most important properties. We chose
environments of increasing complexity that model widely used benchmark
tasks. We list the used algorithms and then describe how we gather
evaluation data and how it is aggregated. Last, we describe the
methods we use for analyzing state-space coverage.

\subsection{Noise types: Gaussian and Ornstein-Uhlenbeck\label{sec:noise_types}}

The action noise $\varepsilon_{a_t}$ is added to the action drawn from
the policy:
\begin{align}
  a_t  =  \clip_{a_{\min}, a_{\max}}\left[\tilde a_t ~~ + ~~ \beta \left( \clip_{-1;1}[\varepsilon_{a_t}] \cdot \frac{a_{\max} - a_{\min}}{2} + \frac{a_{\max} + a_{\min}}{2}\right)\right]  \label{eq:action_noise}
\end{align}

where $\tilde a_t \sim \pi_\theta(\cdot|s_t)$ for stochastic policies or
$\tilde a_t = \pi_\theta(s_t)$ for deterministic policies. We
introduce an additional impact scaling factor $\beta$, which is
typically kept constant at the value one. In \refsec{sec:schedulers}
we describe how we change $\beta$ over time to create a noise
scheduler. The action noise $\varepsilon_{a_t}$ is drawn from either a
Gaussian distribution or an Ornstein-Uhlenbeck (OU) process. The noise
distributions are factorized, i.e.\ noise samples are drawn
independently for each action dimension.
For the generation of action noise samples, the action space is
assumed to be limited to $[-1, 1]$ but then rescaled to the actual limits defined by the environment.

\paragraph{Gaussian noise} is temporally uncorrelated and is typically
applied on symmetric action spaces~\cite{stable-baselines,rl-zoo3}
with commonly used values of $\mu=0$ and $\sigma=0.1$ with
$\Sigma = \bm{I} \cdot \sigma$.  Action noise is sampled according
to
\begin{equation}
\varepsilon_{a_t} \sim \mathcal{N}({\mu, \Sigma}) \label{eq:an_gauss}
\end{equation}
\change{In this setup, Gaussian action noise is sampled and clipped to
  the action limits as needed.\footnote{An alternative way of sampling Gaussian
  action noise would be to use a truncated Gaussian distribution. We
  investigate non-truncated Gaussian distributions together with
  clipping, as they are more common in practice.} 
}

\paragraph{Ornstein-Uhlenbeck noise} is sampled from the following
temporal process, with each action dimension calculated independently
of the other dimensions:
\begin{gather}
\varepsilon_{a_t} =  \varepsilon_{a_{t-1}} + \theta (\mu - \varepsilon_{a_{t-1}}) \cdot \mathrm{d}t + \sigma
  \change{\sqrt{\mathrm{d}t}} \cdot \epsilon_t \label{eq:an_ou} \\
  \varepsilon_{a_0} = \bm{0} ~~~~~ \epsilon_t \sim \mathcal{N}(\bm{0}, \bm{I})
\end{gather}
The process was originally described by \citeintext{uhlenbeckTheoryBrownianMotion1930} and applied to reinforcement learning in DDPG by \citeintext{lillicrapContinuousControlDeep2016}.
The parameters we use for the Ornstein-Uhlenbeck noise are taken from a
widely used RL-algorithm implementation~\cite{stable-baselines}:
$\theta = 0.15$, $\mathrm{d}t = 0.01$, $\mu = \bm{0}$,
$\sigma = 0.1 \cdot \bm{I}$.

Due to the huge number of possible combinations of environments,
algorithms, noise type, noise scale and the necessary repetition with
different seeds, we had to limit the number of investigated scales. We
set out with two noise scales $\sigma$ encountered in pre-tuned
hyper-parameterization \cite{rl-zoo3}, $0.1, 0.5$, and continued with
a linear increase, $0.9, 1.3, 1.7$. Much smaller noise scales vanish
in the variations induced by learning and much larger scales lead to
Bernoulli trials of the min-max actions without much difference.
Because the action noise is clipped to $[-1,1]$ before being scaled to
the actual action limits, a very large scale, such as $1.7$, \change{implies a
larger percentage of on-the-boundary action noise samples and is thus
more similar to bang-bang control actions, the latter having been found
surprisingly effective in many RL benchmarks \cite{seydeBangBangControlAll2021}.}

\subsection{Scheduling strategies to reduce action noise\label{sec:schedulers}}
\begin{figure}[H]
  \includegraphics[width=.49\textwidth]{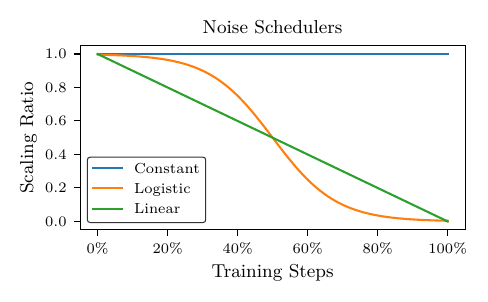}
  \caption{Action noise is used for exploration. The agent should
    favor exploration in the beginning but later favor exploitation.
    Similar to $\epsilon$-greedy strategies in discrete-action
    Q-learning, the logistic and linear schedulers reduce the impact
    of noise (scaling ratio, $\beta$ in \refeq{eq:action_noise}) over the course of the training progress.
    \label{fig:noise_schedulers} }
\end{figure}

In \refeq{eq:action_noise} we introduce the action noise scaling-ratio
$\beta$. In this work we compare a constant-, linear- and
logistic-scheduler for the value of $\beta$. The effective scaling of
the action noise by the noise schedulers is illustrated in
\reffig{fig:noise_schedulers}. The noise types are described in more
detail in \refsec{sec:noise_types}.

Changing the $\sigma$ (see \refeq{eq:an_ou} and \refeq{eq:an_gauss})
instead of $\beta$ could result in a different shape of the
distribution, for example when values are clipped, or when the
$\sigma$ indirectly affects the result as in the Ornstein-Uhlenbeck
process. To keep the action noise distribution shape constant,
the action noise schedulers do not change the $\sigma$ parameter of
the noise process but instead scale down the resulting sampled action
noise values by changing the $\beta$ parameter: this means that the
effective range of the action noise, before scaling and adjusting to
the environment limits, changes over time from $[-1, 1]$, the maximum
range, to $0$ for the linear and logistic schedulers.

\subsection{Environments\label{dbg:environments}}

For evaluation we use various environments of increasing complexity:
Mountain-Car, Inverted-Pendulum-Swingup, Reacher, Hopper, Walker2D,
Half-Cheetah. Observation dimensions range from $2$ to $26$, and
action dimensions range from $1$ to $6$. See \reftbl{tbl:envimgs} for
details, including a rough sketch of the reward. The table indicates
whether the reward is sparse or dense with respect to a goal state,
goal region, or a change of the distance to the goal region. Many
environments feature linear or quadratic (energy) penalties on the
actions (e.g. Hopper). Penalties on the state can be sparse (such as joint limits),
or dense (such as force or required power induced by joint
states).
\citet{brockmanOpenAIGym2016}, \citet{coumansPyBulletPythonModule2016}, and \citet{benelot2018} \change{provide} further details.

\subsection{Performed experiments\label{sec:performed_experiments}}

We evaluated the effects of action noise on the popular and widely-used
algorithms: TD3~\cite{fujimotoAddressingFunctionApproximation2018},
DDPG~\cite{lillicrapContinuousControlDeep2016},
SAC~\cite{haarnojaSoftActorCriticAlgorithms2019}, and a deterministic version
of SAC (DetSAC, \refalg{alg:detsac}). Originally SAC was proposed with
only exploration from its stochastic policy. However, since SAC is an
off-policy algorithm, it is possible to add additional action noise, a
common solution for environments such as the Mountain Car. The
stochastic policy in SAC typically is a parameterized Gaussian and
combining the action noise with the stochasticity of actions sampled
from the stochastic policy could impact the results. Thus, we also
compared to our DetSAC version, where action noise is added to the mean
action of the DetSAC policy (\refalg{alg:detsac}).

We used the implementations provided by \citeintext{stable-baselines3},
following the hyper-parameterizations provided by
\citeintext{rl-zoo3}, but adapting the action noise settings.

The experiments consisted of testing $6$ environments, $4$ algorithms,
$5$ noise scales, $3$ schedulers and $2$ noise types. Each experiment
was repeated with $20$ different seeds, amounting to $14400$
experiments in total. On a single node, \libname{AMD Ryzen 2950X}
equipped with four \libname{GeForce RTX 2070 SUPER, 8 GB}, running
about twenty experiments in parallel this would amount to a runtime of
approximately $244$ node-days \change{(which accounted for about 6 weeks on our cluster)}.

\change{\refsec{sec:experiment_result_details} lists further details
such as the returns averaged across seeds} for each experimental configuration.

\subsection{Measuring Performance\label{sec:measuringperformance}}
For each experiment (i.e.\ single seed), we divided the learning
process into $100$ segments and evaluated the exploration and learned
policy performance \delchange{once} for each of those segments. At the
end of each segment, we performed evaluation rollouts for $100$ episodes
or $10000$ steps, whichever was reached first, using only complete
episodes. \change{This ensures sufficient data points when the episode
  length varies greatly (e.g.\ for the Hopper). This procedure} was
performed for both the deterministic \emph{exploitation} policy as well as
the \emph{exploratory} (action noise) policy. The two resulting
datasets of evaluation rollouts are used to calculate state-space
coverages and returns. These evaluation rollouts, both exploring and
exploiting, were \emph{not} used for training \change{and thus do
  \emph{not} change the amount of training data seen between training
  steps.} We took the mean over these $100$ measurements to aggregate
them into a single value. This is equivalent to measuring the area
under the learning curves. \change{For the evaluation returns, this is
  called the Performance \vP and is our main measure for learning
  performance. Similarily, aggregated evaluation returns measured in
  this fashion are denoted by \vR.}

The learning algorithm uses a noisy (exploratory) policy to collect
data and exploratory return and state-space coverage could be assessed
based on the replay buffer data. However, to get statistically more
robust estimates of the quality of the exploratory policy (returns and
state-space coverage), we performed the above mentioned exploratory
evaluation rollouts and used these rollouts for assessing state-space
coverage and exploratory returns instead of the data in the replay
buffer. \change{Again, these $100$ measurements were aggregated by
  taking the mean and denoted as the exploratory state-space coverage
  \vX and the evaluation state-space coverage \vE.}

\subsection{State-Space Coverage}

\begin{figure}
  \begin{minipage}{\columnwidth}
    \begin{subfigure}[t]{.22\columnwidth}
      \includegraphics[height=0.85\textwidth]{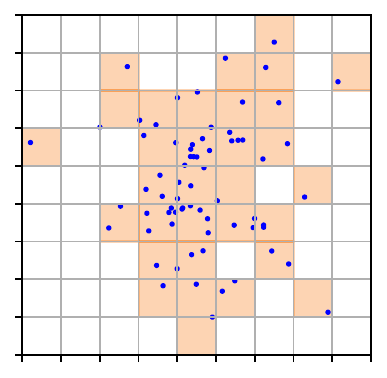}
      \subcaption{$\xbin$ divides the state-space into bins and measures
        the ratio of non-empty bins.}
    \end{subfigure}
    \hspace{0.02\columnwidth}
    \begin{subfigure}[t]{.22\columnwidth}
      \includegraphics[height=0.85\textwidth]{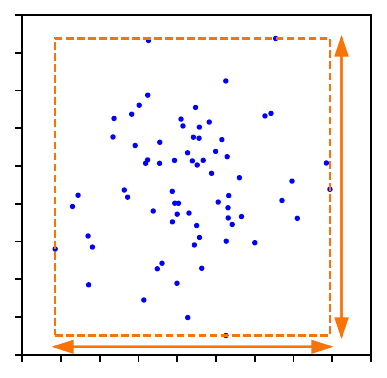}
      \subcaption{$\xbbm$ measures the spread by the mean of the
      side-lengths of the bounding box.}
    \end{subfigure}
    \hspace{0.02\columnwidth}
    \begin{subfigure}[t]{.22\columnwidth}
      \includegraphics[height=0.85\textwidth]{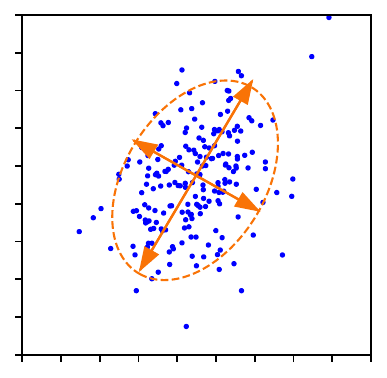}
      \subcaption{$\xnn$ measures the spread of the data by the sum of
      the eigenvalues of the covariance of the data.}
    \end{subfigure}
    \hspace{0.02\columnwidth}
    \begin{subfigure}[t]{.22\columnwidth}
    
      \includegraphics[height=0.85\textwidth]{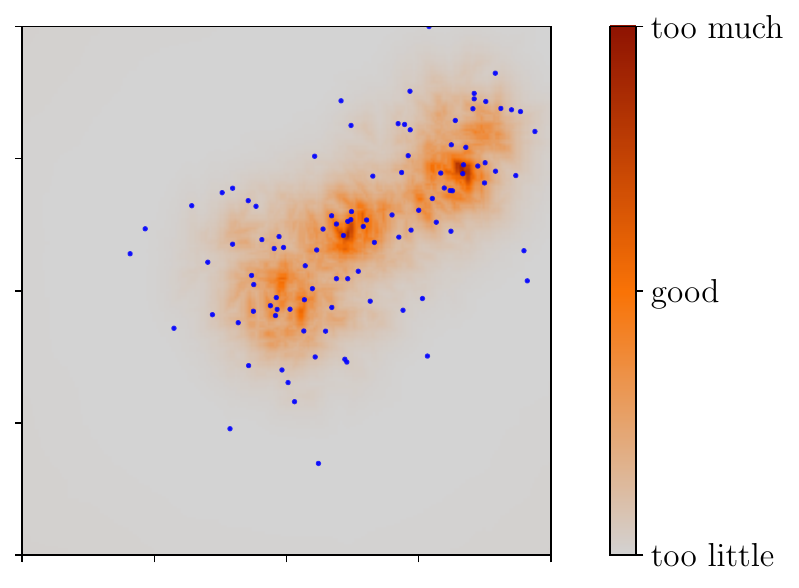}
      \subcaption{$\xurel$ (ours) measures the symmetric KL-di\-ver\-gence
        between a prior over the state space and the collected
        state-space data. }
    \end{subfigure}
  \end{minipage}
  \caption{Illustrations of the state-space coverage
    measures. $\xurel$ scales to high dimensions (unlike $\xbin$) and
    is not susceptible to \change{estimation artifacts due to points
      close to the support boundary} \delchange{boundary artifacts}
    (unlike $\xbbm$, $\xnn$ \change{and kNN based estimators}).
}
  \label{fig:illustrated_measures}
  
\end{figure}

We assess exploration in terms of state-space coverage. We assume that
the environment states $s \in \mathcal{R}^d$ have finite upper and
lower limits: $\mathrm{low} \leq s \leq \mathrm{high}$,
$\mathrm{low}, \mathrm{high} \in \mathcal{R}^d$. We investigate four
measures: $\xbin,\xurel,\xbbm,\xnn$, which are illustrated in
\reffig{fig:illustrated_measures}.

The most intuitive measure for state-space coverage is a histogram-based approach $\xbin$, which divides the state space into equally
many bins along each dimension and measures the ratio of non-empty
bins to the total number of bins:
\begin{equation}
  \xbin= \frac{\# \text{ of non-empty bins}}{\# \text{ number of bins}}
\end{equation}
The number of bins, as the product of divisions along each
dimension, grows exponentially with the dimensionality. This means
that either the number of bins has to be chosen very low, or, if there
are more bins than data points, the ratio has to be adjusted. We chose
to limit the number of bins. For a sample of size $m$ and
dimensionality $d$ the divisions $k$ along each dimension are chosen
to allow for at least $c$ points per bin
\begin{align}
  k = \floor*{ \left(\frac{m}{c}\right)^{\frac{1}{d}}}
\end{align}
However, for high-dimensional data, the number of bins becomes very
small and the measure easily reaches $100\%$ and becomes meaningless,
or, the required number of data points becomes prohibitively large
very quickly. Thus, alternatives are necessary.

\citeintext{zhanTakingScenicRoute2019} proposed two state-space
coverage measures that also work well in high-dimensional spaces: the
\emph{bounding box mean} $\xbbm$, and the \emph{nuclear norm} $\xnn$.
$\xbbm$ measures the spread of the data by a $d$ dimensional bounding
box around the collected data $D = \{\ldots, {\bm s}^{(j)}, \ldots\}$
and measuring the mean of the side lengths of this bounding box:
\begin{equation}
 \xbbm = \frac{1}{d} \sum_i^d \left[ \max_j s_i^{(j)} - \min_j s_i^{(j)} \right]
\end{equation}

$\xnn$, the nuclear norm
estimates the covariance matrix ${C}$ of the data and measures data
spread by the trace, the sum of the eigenvalues of the estimated
covariance:
\begin{equation}
  \xnn(D) := \operatorname{trace} \big( C (D) \big)
\end{equation}

\change{As shown below in \refsec{subsubsec:evalsynthdata}}\delchange{As we will see},
extreme values or values close to the state-space boundaries can lead
to over-estimation of the state-space coverage by these two
measures. We therefore propose a measure more closely related to
$\xbin$ but more suitable to higher dimensions: $\xurel(D)$. The
Uniform-relative-entropy measure
$\xurel$ assesses the uniformity of the collected data, by measuring the
state-space coverage as the symmetric divergence between a uniform
prior over the state space $U$ and the data distribution $Q_D$:
\begin{equation}
  \xurel(D) =
  -\DKL\big({U}||{Q}_D\big) -
  \DKL\big({Q}_D||{U}\big)
\end{equation}
The inspiration for this measure comes from the observation that the
exploration reward for count-based methods without task reward would
be maximized by a uniform distribution. We assume that for robotics
tasks reasonable bounds on the state space can be found. In a bounded
state space, the uniform distribution is the least presumptive
(maximum-entropy) distribution. The addition of the
$\DKL\big({U}||{Q}_D\big)$ term helps to reduce under-estimation of
the divergence in areas with low density in $Q_D$. Note that $Q_D$ is
only available through estimation, and the support for $Q_D$ is never
zero as the density estimate never goes to zero. To estimate the
relative uniform entropy we evaluated two divergence estimators, a
kNN-based (k-Nearest-Neighbor) estimator and a Nearest-Neighbor-Ratio
(NNR) estimator \cite{noshadDirectEstimationInformation2017}. 
\change{Density estimators based on kNN
are susceptible to over- / under-estimation artifacts close
to the boundaries (support) of the state space (see
  \reffig{fig:boundary_artifacts} for an illustration). In contrast,
the NNR estimator does not suffer from these artifacts.  }
If not specified explicitly, $\xurel$ refers to the NNR-based variant.
\emph{\change{kNN $\xurel$ estimator:}} $\xurel$ can be estimated using a 
\emph{kNN density estimate} $\hat q_k(s)$,
as described in \cite{bishopPatternRecognition2006},
where $V_d$ denotes the unit volume of a $d$-dimensional sphere,
$R_k(x)$ is the Euclidean distance to the $k$-th neighbor of $x$, and
$n$ is the total number of samples in $\mathcal{D}$:
\begin{align}
  V_d = & \frac{ \pi^{d/2}}{\Gamma({\frac{d}{2} + 1})} \\
  \hat q_k(x) = & \frac{k}{n} \frac{1}{ V_d R_k(x)^d} =  \frac{k}{n V_d } \frac{1}{ R_k(x)^d}
\end{align}
where $\Gamma$ denotes the gamma function.

\emph{\change{NNR $\xurel$ estimator:}}
Alternatively, $\xurel$ can be \emph{estimated using NNR}, an $f$-divergence
estimator,
based on the ratio of the nearest neighbors around a query point.

For the general case of estimating $\DKL\big(P\vert\vert Q\big)$, we
take samples from $X \sim Q$ and $Y \sim P$. Let $\mathcal{R}_k(Y_i)$
denote the set of the $k$-nearest neighbors of $Y_i$ in the set
$Z:= X \cup Y$. $N_i$ is the number of points from
$X \cap \mathcal{R}_k(Y_i)$, $M_i$ is the number of points from
$Y \cap \mathcal{R}_k(Y_i)$, $M$ is the number of points in $Y$ and
$N$ is the number of points in $X$, $\eta = \frac{M}{N}$. 
\change{The NNR measure requires the density of $P$ and $Q$
to be bounded with the lower limit $C_L > 0$, and measures the ratio of points from
two different distributions around a query point.} 
Assuming all points of a sample of size $n$ are
concentrated around a single point, we \change{lower-bound} the density to $C_L = \frac{1}{n}$.
\change{To limit the peaks around a single point we upper-bound the densities to} $C_U = \frac{n}{1}$.

\begin{align}
  \DKL(P\vert\vert Q) \approx & \hat D_g(X, Y) \\
  \hat D_g(X, Y) := & \max \left(
                      \frac{1}{M} \sum_{i=1}^M \hat g \left(\frac{\eta N_i}{M_i + 1}\right), 0
                      \right) \\
  \text{where }  \hat g(x) := & \max \big(g(x), g(C_L/C_U) \big) \\
  g(\rho) := & -\log\rho
\end{align}
\subsubsection{\label{subsubsec:evalsynthdata}Evaluation of Measures on Synthetic Data}

\begin{figure}[!b]
\centering
\begin{subfigure}{1.0\textwidth}
  \centering
  \includegraphics[width=\textwidth]{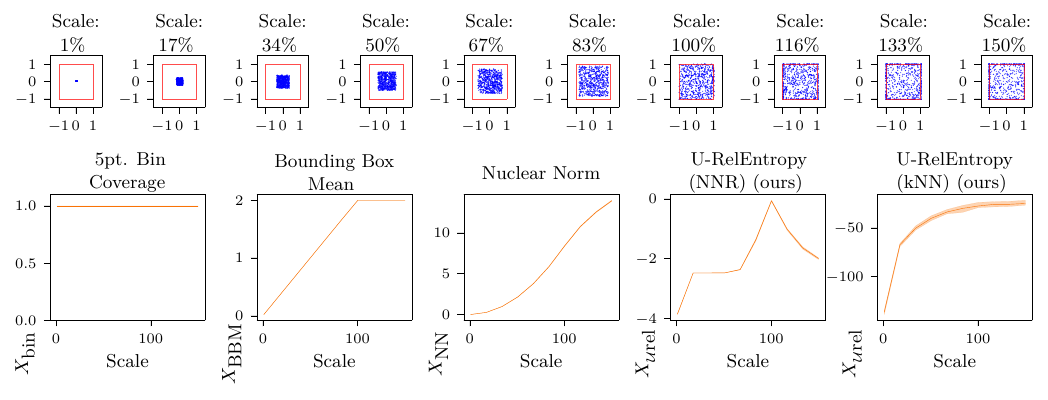}

  \centering \subcaption{Growing Uniform distribution: evaluation of
    the state-space coverage measures on synthetic data -- for larger
    scale values more points are clipped to the state-space
    boundaries, leading to an expected decrease in state-space
    coverage for scales larger than $100\%$. This behavior is only captured by $\xurel$ (NNR).
}
  \label{fig:growing_uniform}
\end{subfigure}

\vspace{1em}

\begin{subfigure}{1.0\textwidth}
  \centering
  \includegraphics[width=\textwidth]{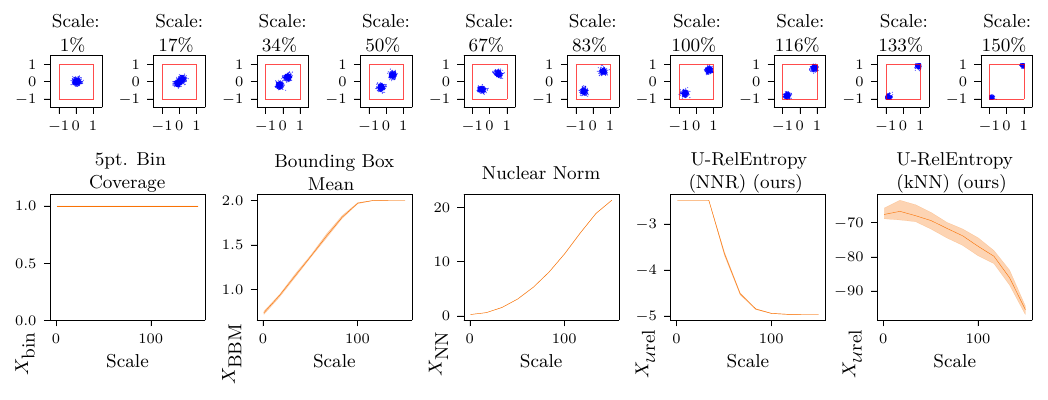}
  \subcaption{Growing Distance of Modes of 2-Mixture of Truncated
    Normal: evaluation of the state-space coverage measures on
    synthetic data. For larger scale values, the location of the
    mixture components is closer to the boundary -- leading to an
    expected reduction in coverage for larger scale values. $\xbin,\xbbm,\xnn$ fail to capture this behavior. }
  \label{fig:truncated_normal_location}
\end{subfigure}\caption{state-space coverage measures may not accurately represent the real coverage.
Each comparison (a-b) shows the different exploration measures
  $\xbin, \xbbm, \xnn$ and $\xurel$ (ours) on synthetic   $\syntheticDatadim$ dimensional data. $\xbin$ becomes constant
  and $\xbbm$ and $\xnn$ suffer from \change{estimation artifacts for points close to the support boundary}.
The different data generating distributions depend on a \emph{scale}
  parameter. The distributions are factorial and similarly distributed
  along each dimension. The scatter plots in (a-b) depict first vs.\
  second dimension. 
\change{The mean and min-max variation is shown. In the majority of cases the 
  variance is surprisingly small.}
\label{fig:results}
}

\end{figure}

To compare the different exploration measures, we assumed a
$d=\syntheticDatadim$ dimensional state space, generated data from two
different types of distributions, and compared the exploration
measures on these data. The experiments were repeated \change{$20$} times, and
\change{the mean and min-max} values are plotted in \reffig{fig:results}. 
\change{Each sampled dataset consists of $5000$ points. For most measures 
the variance is surprisingly small.}
While the data are $d$-dimensional, they come from factorial
distributions, similarly distributed along each dimension. Thus, we
can gain intuition about the distribution from scatter plots of the
first vs.\ second dimension. This is depicted at the top of each of
the two parts. The bottom part of each comparison shows the different
exploration measures, where the scale parameter is depicted on the $x$
axis and the exploration measure on the $y$ axis.

\paragraph{(a) Growing Uniform:} Figure \ref{fig:growing_uniform}
depicts data generated by a uniform distribution, centered around the
middle of the state space, with minimal and maximal values growing
relatively to the full state space according to the \emph{scale}
parameter from $1\%$ to $150\%$. Since in the latter case, many points
would lie outside the allowed state space; these values are clipped to
the state-space boundaries. This loosely corresponds to an
undirectedly exploring agent that overshoots and hits the state-space
limits, sliding along the state-space boundaries. Note how the
estimation (kNN vs. NNR) has a great impact on the $\xurel$ measure's
performance here: We would expect a maximum around a scale of $100\%$
and smaller values before and after (due to clipping). Here the
$\xurel$ (NNR) measure most closely follows this expectation. The
ground-truth value of the divergence would follow a similar
shape. However, since the densities are limited for the NNR estimator,
the ground-truth divergence would show more extreme values.

\paragraph{(b) Bi-Modal Truncated Normal moving locations:} Figure
\ref{fig:truncated_normal_location} shows a mixture of two truncated
Gaussian distributions, with equal standard deviations but located
further and further apart (depending on the scale parameter). In this
case, the state-space coverage should increase until both
distributions are sufficiently far apart, should then stay the same,
and finally begin to drop because the proximity to the
state-space-boundary limits the points to an ever smaller volume.  The
inspiration for this example distribution is an agent setting off in
two opposite directions and getting stuck at these two opposing
limits.  While somewhat contrived and more extreme than the inspiring
example, it highlights difficulties in the exploration measures. Both
the bounding-box mean $X_\text{BBM}$ and the nuclear norm
$X_\text{NN}$ completely fail to account for vastly unexplored areas
between the extreme points.

  Since the $\xurel$ NNR measure is clipped (by definition of NNR) the
  measure reaches its limits when the density ratios become extreme,
  which presumably happens for very small and large scale parameters
  in this setting. The $X_\text{Urel}$ kNN approximator is better able
  to capture the extreme divergence values, however, as pointed out
  before, this comes at the cost of under-estimating the divergence
  for points close to the support boundary.

  The experiments on synthetic data showed that the histogram based
  measure is not useful in high-dimensional spaces. The alternatives
  $\xbbm$ and $\xnn$ are susceptible to artifacts on bounded
  support. This susceptibility to boundary artifacts is also present
  in the kNN-based $\xurel$ estimator, because of these results we
  employ the NNR-estimator based $\xurel$ in the rest of this paper
  and refer to it as $\xurel$.

\section{Results: What action noise to use?}
In this section we analyze the data collected in the experiments
described in \refsec{sec:performed_experiments}. We first look at the experiments
performed under a constant scale scheduler since this is the most
common case in the literature. In this setting we will look at two
aspects:
first, is one of the two action noise types
generally superior to the other \emph{(Q1)}?
And secondly, is there a
generally preferable action noise scale \emph{(Q2)}?
Then,  we compare across constant, linear and logistic schedulers to see if
reducing the noise impact over the training process is a reasonable
thing to do \emph{(Q3)}.
Finally we compare the relative importance of the scheduler,
noise type and scale \emph{(Q4)}.
See \refsec{sec:statmethods} for a brief description of the
statistical methods used in this paper \change{and the verification of their assumptions}.

\subsection{(Q1) Which action noise type to use? (and what are the impacts)\label{sec:q1_what_noise}}

\begin{figure}[bt]
  \noindent
\includegraphics[width=1.0\columnwidth]{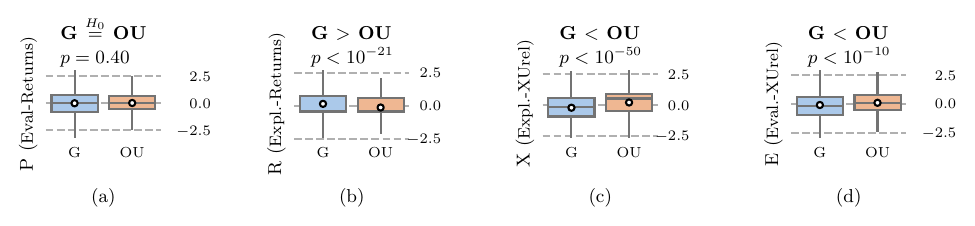}
  \caption{    Comparison of standardized measures (\vP, \vR, \vX, \vE), for
    Gaussian (G) and Ornstein-Uhlenbeck (OU) noise types, (a-d).
    Values are standardized to control for and
    combine algorithm, environment and noise scale:
\emph{(a)} For learned performance \vP, measured by
    evaluation returns, neither of the two noise types is
    significantly better. \emph{(b)} For Returns collected under the
    exploration policy \vR, Gaussian noise collects data with slightly
    better returns ($p <
    10^{-21}$).  (c) For State-space coverage of the exploratory
    policy \vX Ornstein-Uhlenbeck performs better. (d) The State-space
    coverage of evaluation rollouts \vE is slightly larger for
    Ornstein-Uhlenbeck noise without significantly affecting the
    evaluation returns \vP.
Overall neither of the two noise types is superior.
}
  \label{fig:WAN_noise_type_const}
\end{figure}

\begin{table}[bt]
  \begin{adjustbox}{angle=0,max width=\columnwidth,max height=\textheight}
    \begin{tabular}{@{}m{3cm}@{\hskip 0.3in}r@{~}c@{~}r@{}@{\hskip 0.3in}r@{~}c@{~}r@{}@{\hskip 0.3in}r@{~}c@{~}r@{}@{\hskip 0.3in}r@{~}c@{~}r@{}}
\toprule
              Environment &  P &                                 $p_P$ & $d_\text{P}$ &  R &                                 $p_R$ & $d_\text{R}$ &  X &                                 $p_X$ & $d_\text{X}$ &  E &                                 $p_E$ & $d_\text{E}$ \\
\midrule
             Half-Cheetah &  - &               {\footnotesize $0.89$ } &            - &  G &              {\footnotesize $0.002$ } &         0.22 & OU &              {\footnotesize $0.004$ } &         0.21 &  - &               {\footnotesize $0.20$ } &            - \\
                   Hopper & OU &  {\footnotesize $\texttt{<}10^{-3}$ } &         0.27 &  G &  {\footnotesize $\texttt{<}10^{-4}$ } &         0.29 &  G &  {\footnotesize $\texttt{<}10^{-8}$ } &         0.41 &  - &               {\footnotesize $0.71$ } &            - \\
Inverted-Pendulum-Swingup &  - &               {\footnotesize $0.38$ } &            - &  G & {\footnotesize $\texttt{<}10^{-51}$ } &         1.15 & OU & {\footnotesize $\texttt{<}10^{-56}$ } &         1.22 &  G &              {\footnotesize $0.002$ } &         0.22 \\
             Mountain-Car & OU & {\footnotesize $\texttt{<}10^{-10}$ } &         0.47 & OU & {\footnotesize $\texttt{<}10^{-19}$ } &         0.66 & OU &  {\footnotesize $\texttt{<}10^{-5}$ } &         0.34 & OU & {\footnotesize $\texttt{<}10^{-21}$ } &         0.71 \\
                  Reacher &  G & {\footnotesize $\texttt{<}10^{-30}$ } &         0.87 &  G & {\footnotesize $\texttt{<}10^{-26}$ } &         0.80 & OU & {\footnotesize $\texttt{<}10^{-40}$ } &         1.01 & OU & {\footnotesize $\texttt{<}10^{-29}$ } &         0.84 \\
                 Walker2D &  - &              {\footnotesize $0.039$ } &            - &  G &              {\footnotesize $0.010$ } &         0.18 & OU &  {\footnotesize $\texttt{<}10^{-9}$ } &         0.46 &  - &               {\footnotesize $0.28$ } &            - \\
\bottomrule
\end{tabular}
   \end{adjustbox}
  \caption{Per environment the noise type is important: Comparison of Evaluation Returns \vP,
  Exploratory Returns \vR, Exploratory-$\xurel$~\vX,  and Evaluation-$\xurel$ \vE.  Values are standardized
    to control for and aggregate over algorithm, and noise scale. The
    results are compared using a Welch-t-test. Significantly better
    noise type for each environment and measure is reported
    ($p < 0.01$), as well as two-tailed p-values $p_{(\cdot)}$ and
    Cohen-d effect size $d_{(\cdot)}$. While overall neither of the
    two noise types leads to significantly better performance \vP (see
    \reffig{fig:WAN_noise_type_const}), per environment noise type
    difference is significant.}
  \label{tbl:WAN_noise_type_envnames_const}
\end{table}

To compare the impact of the action noise \emph{type}, we look at the
constant $\beta=1$ case, group the aggregated performance and
exploration results (see \refsec{sec:measuringperformance}) by the
factors algorithm, environment, and action noise scale and standardize
the results to control for their influence. These standardized results
are then combined for each noise
type. \reffig{fig:WAN_noise_type_const} illustrates the results. The
comparisons are performed by Welch-t-test, symmetric p-values are
listed.

\reffig{fig:WAN_noise_type_const} (c) shows that
\emph{Ornstein-Uhlenbeck noise leads to increased state-space
  coverage} under the exploratory policy \vX as measured by
$\xurel$. For completeness \reffig{fig:WAN_noise_type_const} (d) shows
the state-space coverage of the evaluation policy. Here
Ornstein-Uhlenbeck increases coverage which might indicate slightly
longer trajectories for policies trained under Ornstein-Uhlenbeck noise,
however whether this is preferable or not is task dependant.
Exploration likely incurs additional costs, e.g.\ through action
penalties, but also by moving the agent away from
high-reward-trajectories. Since Ornstein-Uhlenbeck noise is temporally
correlated, it is more efficient in covering more state-space but also
in moving the agent away from high-reward trajectories.  Thus
\emph{exploratory returns \vR are larger for Gaussian noise} and
conversely smaller for Ornstein-Uhlenbeck noise, see
\reffig{fig:WAN_noise_type_const} (b).
The learning process is able to
offset some differences in the data as shown in
\reffig{fig:WAN_noise_type_const} (a): the significant differences in
exploratory returns \vR and exploratory state-space coverage \vX
\emph{do not} translate into significantly-different performance
\emph{across environments}. When viewed on a per-environment basis,
\reftbl{tbl:WAN_noise_type_envnames_const} (column P) shows that,
\emph{the preferable noise type depends on the environment}:
Ornstein-Uhlenbeck is preferable for Hopper and Mountain-Car, but
Gaussian for the Reacher environment.
\reftbl{tbl:WAN_noise_type_envnames_const} (column X) shows that
Ornstein-Uhlenbeck leads to larger state-space coverage, as before,
and Gaussian noise leads to larger exploratory returns (column R). The
only exceptions to this are the Hopper environment, where the
Ornstein-Uhlenbeck is more likely to topple the agent and the
Mountain-Car environment, where the returns are very closely related
to increasing the state-space coverage and thus exhibits an
improvement of \vR by Ornstein-Uhlenbeck noise.

These results show that the \emph{noise type is important} and
significantly impacts the performance for \emph{some environments}. Neither
of the two noise types leads to better performance, evaluation return
\vP, \emph{in general}. However \emph{Ornstein-Uhlenbeck} generally
\emph{increases state-space coverage}. This is likely due to the
effect, that in many cases the environment acts as an integrator over
the actions\change{: in many environments the action constitutes some 
type of velocity or force control, which by stepping forward, and 
thus integrating forward in time, amounts to changes in position, 
or respectively changes in velocity.}

\subsection{(Q2) Which action noise scale to use?\label{sec:q2_which_scale}}

To analyze the impact of action noise scale, we look at the constant
($\beta=1$) case, and control for the impact of the factors algorithm,
environment and noise type: by grouping the results according to these
factors and standardizing the results. Then results for the same
noise scale are combined.
\begin{figure}
  \noindent
  \includegraphics[page=1,width=1.\columnwidth]{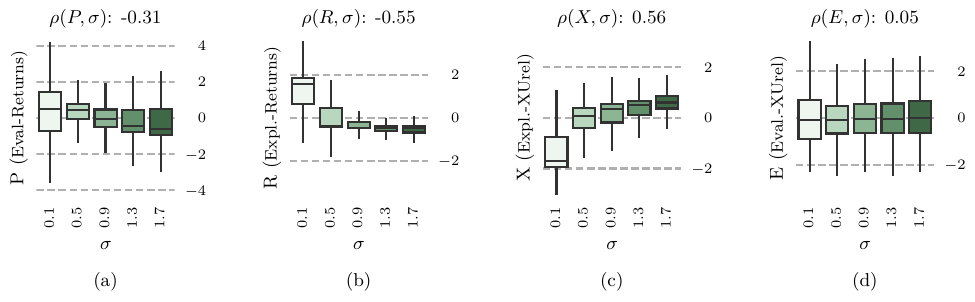}
\caption{Across environments larger noise scales $\sigma$ are
    effective in increasing state-space coverage (c), but reduce
    exploratory returns (b). Measures (\vP, \vX,
    \vR, \vE) are standardized to control for and
    aggregate over algorithm, environment and noise type.  (a)
    Evaluation Performance \vP is negatively correlated with
    action noise scale ($\rho=-0.31$). (b) Larger noise scales
    correlate with smaller exploratory returns \vR. (c)
    Increasing the noise scale $\sigma$ increases exploratory state
    space coverage \vX. (d) State-space coverage of evaluation
    rollouts \vE: the learned trajectories appear unaffected by
    larger noise scale.}
  \label{fig:WAN_noise_scale_const}
\end{figure}

\begin{table}[bt]
  \begin{adjustbox}{angle=0,max width=\textwidth,max height=\textheight}
    \begin{tabular}{lrrrrrr}
\toprule
              Environment &  $\rho(P, R)$ &  $\rho(P, X)$ &  $\rho(P, \sigma_{\textrm{scale}})$ &  $\rho(R, X)$ &  $\rho(R, \sigma_{\textrm{scale}})$ &  $\rho(X, \sigma_{\textrm{scale}})$ \\
\midrule
                      All &          0.57 &         -0.03 &                               -0.31 &         -0.30 &                               -0.55 &                                0.56 \\
             Half-Cheetah &          0.22 &         -0.28 &                               -0.35 &         -0.64 &                               -0.74 &                                0.75 \\
                   Hopper &          0.69 &          0.15 &                               -0.87 &          0.27 &                               -0.74 &                               -0.17 \\
Inverted-Pendulum-Swingup &         -0.15 &          0.23 &                                0.27 &         -0.88 &                               -0.83 &                                0.77 \\
             Mountain-Car &          0.94 &          0.87 &                                0.58 &          0.76 &                                0.37 &                                0.75 \\
                  Reacher &          0.84 &         -0.88 &                               -0.56 &         -0.96 &                               -0.84 &                                0.69 \\
                 Walker2D &          0.76 &         -0.44 &                               -0.81 &         -0.52 &                               -0.82 &                                0.63 \\
\bottomrule
\end{tabular}
   \end{adjustbox}
  \caption{Data quality, measured by exploratory returns \vR,
    does not completely determine performance, measured by evaluation
    returns \vP. $\rho$ denotes Spearman correlation
    coefficients. Generally \vR is positively, but surprisingly
    not always strongly, correlated with \vP. For some
    environments, exploratory state-space coverage \vX is
    beneficial, while generally it is associated with decreased
    evaluation performance \vP. Across environments and
    noise types, increasing the noise scale increases exploratory
    state-space coverage \vX but reduces exploratory returns
    \vR. \label{tbl:WAN_noise_scale_correlations} }
\end{table}

An interesting observation shown in
\reffig{fig:WAN_noise_scale_const}~(c) is that state-space coverage of
the exploratory policy~\vX correlates positively with action noise
scale $\sigma$ ($\rho$ Spearman correlation coefficients). The
takeaway from this is that instead of changing the noise type, one
might \emph{increase state-space coverage by increasing
  $\sigma$}. This however leads to a reduction in the exploratory
returns \vR, see \reffig{fig:WAN_noise_scale_const}~(c),
($\rho(R, \sigma)=-0.55$). Subsequently, larger noise scales~$\sigma$
are associated with decreased learned performance, i.e.\ smaller
evaluation returns \vP, \reffig{fig:WAN_noise_scale_const}~(a), when
viewed across environments. Note that for very small noises
($\sigma = 0.1$) the variance of the results \vP becomes very
large. It appears that, in many cases, less noise is actually better,
but too little noise often does not work well. A good default for
$\sigma$ appears to be $> 0.1$ but $< 0.9$.  The scale~$\sigma$ does
not appear to have a strong effect on the evaluation state-space
coverage \vE, \reffig{fig:WAN_noise_scale_const}~(d). When viewed
separately for each environment
(\reftbl{tbl:WAN_noise_scale_correlations}), the association between
\vX and $\sigma$ is consistent. The only exception is the Hopper task,
where a large noise is more likely to topple the agent, making it fail
earlier, thereby reducing state-space coverage. The association
between $\rho(R, \sigma)$ is consistently negative, with the exception
of the Mountain-Car where more state-space coverage directly
translates to higher returns, because the environment is underactuated
and energy needs to be injected into the system. Offline-RL findings
indicate that it is easier to learn from expert data than from data of
mixed-quality \cite{fuD4RLDatasetsDeep2020}. As such, we would expect
a very strong correlation between exploratory returns \vR as a measure
of data quality and evaluation returns \vP as a measure of learned
performance. Indeed, $\rho(P,R)$ shows that overall exploratory
returns \vR and evaluation returns \vP are mostly positively
correlated. However, the correlation is not always very strong and can
even be negative. This is interesting, because this means that
\emph{exploratory returns are not the only determining factor} for
learned performance. For example, in the Inverted-Pendulum-Swingup,
$\rho(P, R)$ is slightly negative while $\rho(P, X)$ is positive. The
results indicate that, the noise scale~$\sigma$ has to be chosen to
achieve a trade-off between either increasing state-space coverage $X$
or returns $R$ as required for each specific environment.

\subsection{(Q3) Should we scale down the noise over the training process?\label{sec:q3_should_schedule}}

The previous sections indicated that there is no unique solution for
the best noise type and that this choice is dependent on the
environment. The analysis of the noise scale showed an overall
preference for smaller noise scales, but also showed that, in
contrast, some environments require more noise to be solved
successfully. In this section we analyze schedulers that reduce the
influence of action noise ($\beta$) over the training
progress.
\begin{figure}[bt]
  \noindent
  \includegraphics[page=1,width=.8\columnwidth]{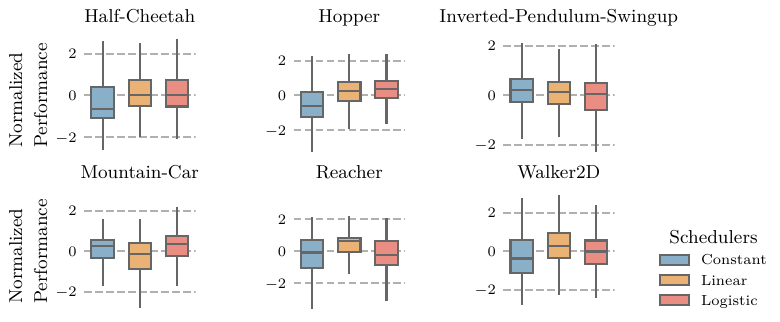}
\caption{
In the majority of cases action noise schedulers improve
    performance. The figure shows the comparison of the learned policy
    performance, measured by evaluation returns \vP, for each
    environment and scheduler. Data is standardized to control for
    influence of algorithm, environment, noise scale $\sigma$ and
    noise type. In the majority of cases the linear and logistic
    schedulers perform better than or comparably to the constant
    scheduler. }
  \label{fig:boxplot_normalized_performance_by_scheduler}
\end{figure}
\begin{table}[bt]
  \begin{adjustbox}{angle=0,max width=\columnwidth,max height=\textheight}
    \begin{tabular}{@{}lccc@{\hskip 0.4in}ccc@{}}
\toprule
\phantom0 & \multicolumn{3}{c}{$\text{var}(P)$} & \multicolumn{3}{c}{$P$} \\
Scheduler & {\footnotesize$<$ Constant} & {\footnotesize$<$ Linear} & {\footnotesize$<$ Logistic} & {\footnotesize$>$ Constant} & {\footnotesize$>$ Linear} & {\footnotesize$>$ Logistic} \\
\midrule
 Constant &                           0 &                         0 &                           0 &                           0 &                         1 &                           1 \\
   Linear &                           4 &                         0 &                           1 &                           4 &                         0 &                           2 \\
 Logistic &                           4 &                         1 &                           0 &                           4 &                         1 &                           0 \\
\bottomrule
\end{tabular}
   \end{adjustbox}
  \caption{In the majority of cases, using a scheduler reduces variance
    of the performance (evaluation returns) \vvarP, and improves expected performance \vP.
The evaluation returns \vP are standardized to control for the
    influence of algorithm, noise scale $\sigma$ and
    noise type. Levene's tests are used to assess difference in variance
    \vvarP and a multiple-comparison Games-Howell test
    indicates superior performance \vP. The table shows the number of environments on which each scheduler (row)
    is significantly better than the other schedulers (column).
See \reftbl{tbl:scheduler_noise_full} for full per-environment results.
}
  \label{tbl:scheduler_noise}
\end{table}

\reffig{fig:boxplot_normalized_performance_by_scheduler} shows the
performance for each environment and each scheduler. The data is
normalized by environment and algorithm before aggregation. The
general tendency observed across environments is that, when the
environment reacts negatively to larger action noise scale
(Half-Cheetah, Hopper, Reacher, Walker2D; as shown in
\reftbl{tbl:WAN_noise_scale_correlations}), \emph{reducing the noise
  impact} $\beta$ over time consistently \emph{improves
  performance}. The \emph{reverse} effect appears to be \emph{less
  important}: for environments benefiting from larger noise scales,
the constant scheduler does not consistently outperform the linear and
logistic schedulers.

\reftbl{tbl:scheduler_noise} shows summarized results indicating the
number of environments where scheduler (1), indicated by row, is
better than scheduler (2), indicated by column, in terms of variance
\vvarP and mean performance \vP. See \reftbl{tbl:scheduler_noise_full}
for full results on the pairwise comparisons. Performance differences
are assessed by a Games-Howell multiple comparisons test, while
variance is compared using Levene's test.

The tests underlying \reftbl{tbl:scheduler_noise} show that the
differences observed in
\reffig{fig:boxplot_normalized_performance_by_scheduler} are indeed
significant. Furthermore, the schedulers (linear, logistic) reduce
variance \vvarP compared to the constant case in four
out of six cases. Keeping the impact $\beta$ constant has no
beneficial effect on variance in any environment. This indicates that
\emph{using a scheduler} to reduce action noise impact increases
\emph{consistency in terms of learned performance}.

\subsection{(Q4) How important are the different parameters?}

In the previous sections we looked at each noise configuration
parameter independently, first for the constant $\beta$ case (Q1, Q2),
secondly for scheduled reduction of $\beta$ (Q3). However, the
question remains whether all the parameters are equally important. We
standardize results to control for environment and algorithm, and
compare across all noise types, noise scales $\sigma$ and all three
schedulers.

\begin{table}
  \begin{adjustbox}{angle=0,max width=\textwidth,max height=\textheight}
    \begin{tabular}{@{}lrrr@{\hskip 0.3in}rrr@{}}
\toprule
 & \multicolumn{3}{c}{Spearman Correlation} & \multicolumn{3}{c}{$\eta^2$ Effect Size} \\
Envname & $\rho(P, X)$ & $\rho(P, \sigma)$ & $\rho(P, R)$ & $\eta^2_\textrm{Scheduler}$ & $\eta^2_\textrm{Type}$ & $\eta^2_\sigma$ \\
\midrule
All & -0.503 & -0.120 & 0.770 & 0.005 & 0.000 & $\mathbf{ 0.084 }$\\
Mountain-Car & 0.662 & 0.442 & 0.959 & 0.032 & 0.060 & $\mathbf{ 0.261 }$\\
Inverted-Pendulum-Swingup & -0.003 & 0.123 & 0.163 & 0.005 & 0.009 & $\mathbf{ 0.115 }$\\
Reacher & -0.872 & -0.382 & 0.803 & 0.048 & $\mathbf{ 0.181 }$& $\mathbf{ 0.181 }$\\
Hopper & -0.349 & -0.599 & 0.651 & 0.045 & 0.022 & $\mathbf{ 0.660 }$\\
Walker2D & -0.658 & -0.494 & 0.677 & 0.014 & 0.017 & $\mathbf{ 0.607 }$\\
Half-Cheetah & -0.581 & -0.259 & 0.745 & 0.007 & 0.002 & $\mathbf{ 0.148 }$\\
\bottomrule
\end{tabular}
   \end{adjustbox}
  \caption{
    Spearman correlation coefficients and ANOVA $\eta^2$ effect sizes on \vP for:
    scheduler, noise type and noise scale $\sigma$.
\change{Action} noise scale $\sigma$ is associated with the largest effect size for evaluation returns \vP.
Results are shown across all environments (standardized and
    controlled for environment and algorithm, first row), and per environment
    (standardized and controlled for algorithm). Generally,
    exploratory returns \vR and evaluation performance \vP are
    positively associated, while generally larger state-space coverage
    \vX appears to impact performance \vP negatively. }
  \label{tbl:all_anova}
\end{table}

\reftbl{tbl:all_anova} shows Spearman correlation coefficients
$\rho(P,X), \rho(P,\sigma), \rho(P, R)$ across all three schedulers (compare to \reftbl{tbl:WAN_noise_scale_correlations} which showed correlations
for the constant $\beta=1$ case only).
Across environments the
schedulers \emph{reduce} correlation $\rho(P, \sigma)$ between learned performance (measured by
evaluation returns \vP) and noise scale $\sigma$: from
$\rho(P, \sigma) = -0.31$ in the constant scheduler case to
$\rho(P, \sigma) = -0.12$ when compared across all three types of
schedulers. This is a further indication that using a scheduler
increases robustness to $\sigma$.
The correlations between
$\rho(P, R)$ are increased to $0.77$ vs.\ $0.57$, presumably because
reducing $\beta$ makes the exploratory policy \emph{more on-policy}
and thus \vP and $R$ become more similar. Interestingly, the
schedulers also increase the negative correlation $\rho(P, X)$ between
the performance and the exploratory state-space coverage, from $-0.03$
in the constant case to $-0.50$ when viewed across all
schedulers. This could be driven by the environments reacting
positively to reduced state-space coverage, which under the schedulers
achieve more runs high in \vR but low in \vX, and thus a stronger
negative correlation.

The three columns on the right in \reftbl{tbl:all_anova} show $\eta^2$ effect
sizes of a three-way ANOVA on the evaluation returns \vP:
$\eta^2_{\textrm{Scheduler}}$, $\eta^2_{\textrm{Type}}$,
$\eta^2_{\sigma}$. The $\eta^2$ effect sizes measure the
percentage-of-total~variance explained by each factor. Only in the Reacher environment, action noise \emph{type} is very
important. Surprisingly, in \emph{all cases} the most important factor
is \emph{action noise scale}, while the requirement for a large or
small action noise scale varies for each environment.

\section{Discussion \& Recommendations\label{sec:discuss_recommend}}

The experiments conducted in this paper showed that the action noise
does, depending on the environment, have a \emph{significant} impact
on the evaluation performance of the learned policy (Q1). Which action
noise type is best unfortunately \emph{depends on the environment}.
For the action noise scale (Q2), our results have shown that
generally a larger noise scale increases state-space coverage.
But since for many environments, learning performance is negatively
associated with larger state-space coverage, a large noise scale does
not generally have a preferable impact. Similarly, very small scales also appear
not to have a preferable impact, as they appear to increase variance
of the evaluation performance (\reffig{fig:WAN_noise_scale_const}).
However, overall, reducing the action noise scaling factor over time
(Q3) mostly has positive effects.
Finally we also looked at all factors concurrently (Q4) and found that
for most environments noise scale is the most important factor.

It is difficult to draw general conclusions from a limited set of
environments and extending the evaluation is limited by the
prohibitively large computational costs. However, we would like to
provide heuristics derived from our observations that may guide the
search for the right action noise.
\begin{table}
  \begin{adjustbox}{angle=0,max width=\textwidth,max height=\textheight}
     \aboverulesep = 0.605mm
     \belowrulesep = 0.605mm
    \begin{tabular}{@{}llllll@{}}
\cmidrule[\heavyrulewidth](r){1-4} \cmidrule[\heavyrulewidth](l){5-6}
                  Envname & Scheduler & $\sigma$ &     Type & Horizon &               Recommendation \\
\cmidrule(r){1-4} \cmidrule(l){5-6}
                      All &       lin &  0.1/0.5 &       OU &         &                              \\
             Mountain-Car &       log &      1.7 &       OU &       L &    large $\sigma$, OU, sched \\
Inverted-Pendulum-Swingup &       con &      0.5 &    Gauss &       L &               large $\sigma$ \\
                  Reacher &       lin &      0.1 &    Gauss &       - & small $\sigma$, Gauss, sched \\
                   Hopper &   lin/log &      0.1 &       OU &       S &    small $\sigma$, sched, OU \\
                 Walker2D &       lin &      0.1 &       OU &       S &    small $\sigma$, OU, sched \\
             Half-Cheetah &   lin/log &      0.5 & Gauss/OU &       S &               small $\sigma$ \\
\cmidrule[\heavyrulewidth](r){1-4} \cmidrule[\heavyrulewidth](l){5-6}
\end{tabular}
   \end{adjustbox}
  \caption{Comparison of best-ranked noise type, scale and scheduler
    across all environments and for each environment
    individually. Scheduler, type and scale are investigated
    separately by standardizing the values to control for environment,
    algorithm and the other two respective factors. Horizon indicates
    whether we expect a long (L) or short (S) effective planning horizon.
    Recommendation indicates action noise configuration choices in
    order of importance as per \reftbl{tbl:all_anova}, for options
    with effect sizes $\eta^2 > 0.01$ (small effect).
    }
  \label{tbl:bestrank}
\end{table}
\reftbl{tbl:bestrank} shows the best-ranking scheduler, scale and type
configurations for each, and across environments. The ranking is based
on the count of significantly better comparisons (pairwise
Games-Howell test on difference, $p \le 0.01$, positive test
statistic). For each of scheduler, type and scale we standardize to
control for the other two factors.
Intuitively, the locomotion environments require only a short effective
planning horizon: the reward in the environments is based on the
distance moved and is relevant as soon as the locomotion pattern is
repeated; for example a $30$-step horizon is enough for similar
locomotion benchmarks
\cite{pinneriSampleefficientCrossEntropyMethod2020}. In contrast,
the Mountain-Car environment only provides informative reward at the
end of a successful episode and thus, the planning horizon needs to be
long enough to span a complete successful trajectory (e.g.\ closer to
$100$ steps). Similarly, the Inverted-Pendulum-Swingup uses a shaped
reward that does not account for spurious local optima: to swing up
and increase system energy, the distance to the goal has to be
increased again. These observations are indicated in the column
\emph{Horizon} (\reftbl{tbl:bestrank}). Finally, the recommendation column interprets the
best-ranked results under the observed importance (Q4) reported in
\reftbl{tbl:all_anova}.
Given these results, we provide the following intuitions as a starting
point for optimizing the action noise parameters (read as: to address this
$\rhd$ do that):
\begin{description}[leftmargin=0pt,itemsep=2pt,parsep=2pt]
\item[Environment is under-actuated $\rhd$ increase state-space coverage] We found that
  in the case of the Mountain-Car and the Inverted-Pendulum-Swingup,
  both of which are underactuated tasks and require a swinging up
  phase, larger state-space coverages or larger action noise scales
  appear beneficial (\reftbl{tbl:WAN_noise_scale_correlations} and
  \reftbl{tbl:all_anova}). Intuitively, under-actuation implies
  harder-to-reach state-space areas.
\item[Reward \change{shape} is misleading $\rhd$ increase state-space coverage]
  Actions are penalized in the Mountain Car by an action-energy
  penalty, which means not performing any action forms a local
  optimum.  In the case of the Inverted-Pendulum-Swingup, the distance
  to the goal forms a shaped reward. However, when swinging up,
  increasing the distance to the goal is necessary. Thus, the shaped
  reward can be \change{\emph{misleading}}: \change{following the reward gradient 
  to \emph{greedily} leads the agent to a spurious local optimum.} \delchange{moves the agent away from taking
  necessary steps.} Optimizing for a spurious local optimum implies not
  reaching areas of the state space where the actual goal would be
  found, thus the state-space coverage needs to be increased to find
  these areas.
\item[Horizon is short $\rhd$ reduce state-space coverage] The
  environments Hopper, Reacher, Walker2D model locomotion tasks with
  repetitive movement sequences.  In the Mountain-Car, positive reward
  is only achieved at the successful end of the episode, where as in
  the locomotion tasks positive reward is received after each
  successful cycle of the locomotion pattern. Thus effectively the
  required planning horizon is shorter compared to tasks such as the
  Mountain-Car. Consistently with the previous point, if the effective
  horizon is shorter, the rewards are shaped more efficiently, we see
  negative correlations with the state-space coverage and the noise
  scale: if the planning horizon is shorter, the reward can be
  optimized more greedily, meaning the state-space coverage can be
  more focused and thus smaller.

\item[Need more state-space coverage $\rhd$ increase scale] Our
  analysis showed that, to increase state-space coverage, one way is
  to increase the scale of the action noise. This leads to a higher
  probability of taking larger actions. In continuous control domains,
  actions are typically related to position-, velocity- or
  torque-control. In position-control, larger actions are directly
  related to more extreme positions in the state space. In velocity
  control, larger actions lead to moving away from the initial state
  more quickly. In torque control, larger torques lead to more energy
  in the system and larger velocities. Currently most policies in D-RL
  are either uni-modal stochastic policies, or deterministic policies.
  In both cases, larger action noise leads to a broader selection of
  actions and, by the aforementioned mechanism, to a broader
  state-space coverage. Note that while this is the general effect we
  observed, it is also possible that a too large action can have a
  detrimental effect, e.g. the Hopper falling, and the premature end
  of the episode will lead to a reduction of the state-space coverage.
\item[Need more state-space coverage $\rhd$ try Ornstein-Uhlenbeck]
Depending on the environment dynamics, correlated noise
  (Ornstein-Uhlenbeck) can increase the state-space coverage: for
  example, if the environment shows integrative behavior over the
  actions, temporally uncorrelated noise (Gaussian) leads to more
  actions that ``undo'' previous progress and thus less coverage. Thus
  correlated Ornstein-Uhlenbeck noise helps to increase state-space
  coverage.
\item[Need less state-space coverage or on-policy data $\rhd$ reduce
  scale | use scheduler to decrease $\beta$] If the policy is already
  sufficiently good, or the reward is shaped well enough, exploration
  should focus around good trajectories. This can be achieved using a
  small noise scale $\sigma$. However, if the environment requires
  more exploration to find a reward signal, it makes to sense to use a
  larger action noise scale $\sigma$ in the beginning while gradually
  reducing the impact of the noise (Q3). The collected data then
  gradually becomes ``more on-policy''.
\item[In general $\rhd$ use a scheduler] We found that using
  schedulers to reduce the impact of action noise over time, decreases
  variance of the performance, and thus makes the learning more
  robust, while also generally increasing the evaluation performance
  overall. Presumably because, once a trajectory to the goal is found,
  more fine grained exploration around the trajectory is better able
  to improve performance.
\end{description}

\section{Conclusion}

In this paper we present an extensive empirical study on the impact of
action noise configurations. We compared the two most prominent action
noise types: Gaussian and Ornstein-Uhlenbeck, different scale
parameters ($0.1, 0.5, 0.9, 1.3, 1.7$), proposed a scheduled reduction
of the impact $\beta$ of the action noise over the training progress
and proposed the state-space coverage measure $\xurel$ to assess the
achieved exploration in terms of state-space coverage. We compared
DDPG, TD3, SAC, and its deterministic variant detSAC on the
benchmarks Mountain-Car, Inverted-Pendulum-Swingup, Reacher, Hopper,
Walker2D, and Half-Cheetah.

We found that (Q1) neither of the two noise types (Gaussian,
Ornstein-Uhlenbeck) is generally superior across environments, but
that the impact of noise type on learned performance can be significant
when viewed separately for each environment: the \emph{noise type}
  needs to be chosen to \emph{fit the environment}.
We found that (Q2) increasing action noise scale, across environments,
increases state-space coverage but tends to reduce learned
performance. Again, whether state-space coverage and performance are
positively correlated, and thus a larger scale is desired,
\emph{depends on the environment}. The positive or negative
\emph{correlation should guide the selection} of action noise.
Reducing the impact ($\beta$) of action noise over training time (Q3),
improves performance in the majority of cases and decreases variance
in performance and thus \emph{increases robustness} to the
action noise \emph{choice}.
Surprisingly, we found (Q4) that the \emph{most important factor}
appears to be the action noise \emph{scale} $\sigma$: if less state-space
coverage is required, the scale can be reduced. More state-space
coverage can be achieved by increasing the action noise scale. This
approach is successful even for Gaussian noise on the Mountain-Car.
We synthesized our results into a set of \emph{heuristics} on how to
choose the action noise based on the properties of the environment.
Finally we \emph{recommend a scheduled reduction} of the action noise
impact factor $\beta$ of over the training progress to improve
robustness to the action noise configuration.

\subsubsection*{Acknowledgments}

We would like to thank Bart Keulen, David Peer, Onno Eberhard,
Sebastian Blaes and the TMLR Reviewers for the useful discussion.
\bibliographystyle{tmlr}  \bibliography{extra.bib,pps-exploration-metric-manual.bib}{} 

\begin{thebibliography}{59}
\providecommand{\natexlab}[1]{#1}
\providecommand{\url}[1]{\texttt{#1}}
\expandafter\ifx\csname urlstyle\endcsname\relax
  \providecommand{\doi}[1]{doi: #1}\else
  \providecommand{\doi}{doi: \begingroup \urlstyle{rm}\Url}\fi

\bibitem[Amin et~al.(2021)Amin, Gomrokchi, Satija, {van Hoof}, and
  Precup]{aminSurveyExplorationMethods2021}
Susan Amin, Maziar Gomrokchi, Harsh Satija, Herke {van Hoof}, and Doina Precup.
\newblock A {{Survey}} of {{Exploration Methods}} in {{Reinforcement
  Learning}}.
\newblock \emph{arXiv:2109.00157 [cs]}, September 2021.

\bibitem[Bellemare et~al.(2013)Bellemare, Naddaf, Veness, and
  Bowling]{bellemareArcadeLearningEnvironment2013}
Marc~G. Bellemare, Yavar Naddaf, Joel Veness, and Michael Bowling.
\newblock The arcade learning environment: {{An}} evaluation platform for
  general agents.
\newblock \emph{Journal of Artificial Intelligence Research}, 47:\penalty0
  253--279, 2013.

\bibitem[Bishop(2006)]{bishopPatternRecognition2006}
Christopher~M. Bishop.
\newblock Pattern recognition.
\newblock \emph{Machine Learning}, 128:\penalty0 1--58, 2006.

\bibitem[Boneau(1960)]{boneauEffectsViolationsAssumptions1960}
C.~Alan Boneau.
\newblock The effects of violations of assumptions underlying the t test.
\newblock \emph{Psychological Bulletin}, 57:\penalty0 49--64, 1960.

\bibitem[Brockman et~al.(2016)Brockman, Cheung, Pettersson, Schneider,
  Schulman, Tang, and Zaremba]{brockmanOpenAIGym2016}
Greg Brockman, Vicki Cheung, Ludwig Pettersson, Jonas Schneider, John Schulman,
  Jie Tang, and Wojciech Zaremba.
\newblock {{OpenAI Gym}}.
\newblock \emph{arXiv:1606.01540 [cs]}, June 2016.

\bibitem[Brown \& {Alan B. Forsythe}(1974)Brown and {Alan B.
  Forsythe}]{brownRobustTestsEquality1974}
Morton~B. Brown and {Alan B. Forsythe}.
\newblock Robust tests for the equality of variances.
\newblock \emph{Journal of the American Statistical Association}, 69\penalty0
  (346):\penalty0 364--367, 1974.

\bibitem[Burda et~al.(2019)Burda, Edwards, Storkey, and
  Klimov]{burdaExplorationRandomNetwork2019}
Yuri Burda, Harrison Edwards, Amos~J. Storkey, and Oleg Klimov.
\newblock Exploration by random network distillation.
\newblock In \emph{{{International Conference}} on {{Learning
  Representations}}}, 2019.

\bibitem[Caspi et~al.(2017)Caspi, Leibovich, Endrawis, and
  Novik]{caspiReinforcementLearningCoach2017}
Itai Caspi, Gal Leibovich, Shadi Endrawis, and Gal Novik.
\newblock Reinforcement {{Learning Coach}}.
\newblock Zenodo, December 2017.
\newblock URL \url{https://doi.org/10.5281/zenodo.1134899}.

\bibitem[Chou et~al.(2017)Chou, Maturana, and
  Scherer]{chouImprovingStochasticPolicy2017}
Po-Wei Chou, Daniel Maturana, and Sebastian~A. Scherer.
\newblock Improving stochastic policy gradients in continuous control with deep
  reinforcement learning using the beta distribution.
\newblock In \emph{International Conference on Machine Learning}, volume~70,
  pp.\  834--843, 2017.

\bibitem[Colas et~al.(2018)Colas, Sigaud, and
  Oudeyer]{colasGEPPGDecouplingExploration2018}
C{\'e}dric Colas, Olivier Sigaud, and Pierre-Yves Oudeyer.
\newblock {{GEP-PG}}: {{Decoupling}} exploration and exploitation in deep
  reinforcement learning algorithms.
\newblock In \emph{International Conference on Machine Learning}, volume~80,
  pp.\  1038--1047, 2018.

\bibitem[Coumans \& Bai(2016--2021)Coumans and
  Bai]{coumansPyBulletPythonModule2016}
Erwin Coumans and Yunfei Bai.
\newblock {{PyBullet}}, a {{Python}} module for physics simulation for games,
  robotics and machine learning.
\newblock 2016--2021.
\newblock URL \url{http://pybullet.org}.

\bibitem[Ellenberger(2018)]{benelot2018}
Benjamin Ellenberger.
\newblock {{PyBullet}} gymperium.
\newblock \emph{GitHub repository}, 2018.
\newblock URL \url{https://github.com/benelot/pybullet-gym}.

\bibitem[Fu et~al.(2020)Fu, Kumar, Nachum, Tucker, and
  Levine]{fuD4RLDatasetsDeep2020}
Justin Fu, Aviral Kumar, Ofir Nachum, George Tucker, and Sergey Levine.
\newblock {{D4RL}}: {{Datasets}} for {{Deep Data-Driven Reinforcement
  Learning}}.
\newblock \emph{arXiv:2004.07219}, 2020.

\bibitem[Fujimoto et~al.(2018)Fujimoto, {van Hoof}, and
  Meger]{fujimotoAddressingFunctionApproximation2018}
Scott Fujimoto, Herke {van Hoof}, and David Meger.
\newblock Addressing {{Function Approximation Error}} in {{Actor-Critic
  Methods}}.
\newblock In \emph{International {{Conference}} on {{Machine Learning}}}, pp.\
  1587--1596. {PMLR}, October 2018.

\bibitem[Fujita et~al.(2021)Fujita, Nagarajan, Kataoka, and
  Ishikawa]{fujitaChainerRLDeepReinforcement2021}
Yasuhiro Fujita, Prabhat Nagarajan, Toshiki Kataoka, and Takahiro Ishikawa.
\newblock {{ChainerRL}}: {{A Deep Reinforcement Learning Library}}.
\newblock \emph{Journal of Machine Learning Research}, 22\penalty0
  (77):\penalty0 1--14, 2021.
\newblock ISSN 1533-7928.

\bibitem[Games \& Howell(1976)Games and
  Howell]{gamesPairwiseMultipleComparison1976}
Paul~A. Games and John~F. Howell.
\newblock Pairwise {{Multiple Comparison Procedures}} with {{Unequal N}}'s
  and/or {{Variances}}: {{A Monte Carlo Study}}.
\newblock \emph{Journal of Educational Statistics}, 1\penalty0 (2):\penalty0
  113--125, 1976.

\bibitem[Haarnoja et~al.(2019)Haarnoja, Zhou, Hartikainen, Tucker, Ha, Tan,
  Kumar, Zhu, Gupta, Abbeel, and Levine]{haarnojaSoftActorCriticAlgorithms2019}
Tuomas Haarnoja, Aurick Zhou, Kristian Hartikainen, George Tucker, Sehoon Ha,
  Jie Tan, Vikash Kumar, Henry Zhu, Abhishek Gupta, Pieter Abbeel, and Sergey
  Levine.
\newblock Soft {{Actor-Critic Algorithms}} and {{Applications}}.
\newblock \emph{arXiv:1812.05905 [cs, stat]}, January 2019.

\bibitem[Hill et~al.(2018)Hill, Raffin, Ernestus, Gleave, Traore, Dhariwal,
  Hesse, Klimov, Nichol, Plappert, Radford, Schulman, Sidor, and
  Wu]{stable-baselines}
Ashley Hill, Antonin Raffin, Maximilian Ernestus, Adam Gleave, Rene Traore,
  Prafulla Dhariwal, Christopher Hesse, Oleg Klimov, Alex Nichol, Matthias
  Plappert, Alec Radford, John Schulman, Szymon Sidor, and Yuhuai Wu.
\newblock Stable {Baselines}.
\newblock GitHub repository, 2018.
\newblock URL \url{https://github.com/hill-a/stable-baselines}.

\bibitem[Hoffman et~al.(2020)Hoffman, Shahriari, Aslanides, {Barth-Maron},
  Behbahani, Norman, Abdolmaleki, Cassirer, Yang, Baumli, Henderson, Novikov,
  Colmenarejo, Cabi, Gulcehre, Paine, Cowie, Wang, Piot, and {de
  Freitas}]{hoffmanAcmeResearchFramework2020}
Matt Hoffman, Bobak Shahriari, John Aslanides, Gabriel {Barth-Maron}, Feryal
  Behbahani, Tamara Norman, Abbas Abdolmaleki, Albin Cassirer, Fan Yang, Kate
  Baumli, Sarah Henderson, Alex Novikov, Sergio~G{\'o}mez Colmenarejo, Serkan
  Cabi, Caglar Gulcehre, Tom~Le Paine, Andrew Cowie, Ziyu Wang, Bilal Piot, and
  Nando {de Freitas}.
\newblock Acme: {{A Research Framework}} for {{Distributed Reinforcement
  Learning}}.
\newblock \emph{arXiv:2006.00979 [cs]}, June 2020.

\bibitem[Hollenstein et~al.(2021)Hollenstein, Saveriano, Sayantan, Renaudo, and
  Piater]{hollensteinHowDoesType2021}
Jakob Hollenstein, Matteo Saveriano, Auddy Sayantan, Erwan Renaudo, and Justus
  Piater.
\newblock How does the type of exploration-noise affect returns and exploration
  on {{Reinforcement Learning}} benchmarks?
\newblock In \emph{Austrian Robotics Workshop}, pp.\  22--26, 2021.

\bibitem[Hong et~al.(2018)Hong, Shann, Su, Chang, Fu, and
  Lee]{hongDiversitydrivenExplorationStrategy2018}
Zhang-Wei Hong, Tzu-Yun Shann, Shih-Yang Su, Yi-Hsiang Chang, Tsu-Jui Fu, and
  Chun-Yi Lee.
\newblock Diversity-driven exploration strategy for deep reinforcement
  learning.
\newblock In \emph{Advances in {{Neural Information Processing Systems}}}, pp.\
   10489--10500, 2018.

\bibitem[Jones et~al.(2001)Jones, Oliphant, and
  Peterson]{jonesSciPyOpenSource2001}
Eric Jones, Travis Oliphant, and Pearu Peterson.
\newblock {{SciPy}}: {{Open}} source scientific tools for {{Python}}, 2001.
\newblock URL \url{http://www.scipy.org}.

\bibitem[Kalashnikov et~al.(2018)Kalashnikov, Irpan, Pastor, Ibarz, Herzog,
  Jang, Quillen, Holly, Kalakrishnan, Vanhoucke, and
  Levine]{kalashnikovQTOptScalableDeep2018}
Dmitry Kalashnikov, Alex Irpan, Peter Pastor, Julian Ibarz, Alexander Herzog,
  Eric Jang, Deirdre Quillen, Ethan Holly, Mrinal Kalakrishnan, Vincent
  Vanhoucke, and Sergey Levine.
\newblock {{QT-Opt}}: {{Scalable Deep Reinforcement Learning}} for
  {{Vision-Based Robotic Manipulation}}.
\newblock \emph{arXiv:1806.10293}, 2018.

\bibitem[Ladosz et~al.(2022)Ladosz, Weng, Kim, and
  Oh]{ladoszExplorationDeepReinforcement2022}
Pawel Ladosz, Lilian Weng, Minwoo Kim, and Hyondong Oh.
\newblock Exploration in deep reinforcement learning: {{A}} survey.
\newblock \emph{Inf. Fusion}, 85:\penalty0 1--22, 2022.
\newblock \doi{10.1016/j.inffus.2022.03.003}.

\bibitem[Liang et~al.(2018)Liang, Liaw, Nishihara, Moritz, Fox, Goldberg,
  Gonzalez, Jordan, and Stoica]{liangRLlibAbstractionsDistributed2018}
Eric Liang, Richard Liaw, Robert Nishihara, Philipp Moritz, Roy Fox, Ken
  Goldberg, Joseph Gonzalez, Michael~I. Jordan, and Ion Stoica.
\newblock {{RLlib}}: {{Abstractions}} for distributed reinforcement learning.
\newblock In Jennifer~G. Dy and Andreas Krause (eds.), \emph{Proceedings of the
  35th International Conference on Machine Learning, {{ICML}} 2018,
  Stockholmsm\"assan, Stockholm, Sweden, July 10-15, 2018}, volume~80 of
  \emph{Proceedings of Machine Learning Research}, pp.\  3059--3068. {PMLR},
  2018.

\bibitem[Lillicrap et~al.(2016)Lillicrap, Hunt, Pritzel, Heess, Erez, Tassa,
  Silver, and Wierstra]{lillicrapContinuousControlDeep2016}
Timothy~P. Lillicrap, Jonathan~J. Hunt, Alexander Pritzel, Nicolas Heess, Tom
  Erez, Yuval Tassa, David Silver, and Daan Wierstra.
\newblock Continuous control with deep reinforcement learning.
\newblock In \emph{Proc. 4th {{Int}}. {{Conf}}. {{Learning Representations}},
  ({{ICLR}})}, 2016.

\bibitem[Lumley et~al.(2002)Lumley, Diehr, Emerson, and
  Chen]{lumleyImportanceNormalityAssumption2002}
Thomas Lumley, Paula Diehr, Scott Emerson, and Lu~Chen.
\newblock The importance of the normality assumption in large public health
  data sets.
\newblock \emph{Annual review of public health}, 23\penalty0 (1):\penalty0
  151--169, 2002.

\bibitem[Mania et~al.(2018)Mania, Guy, and Recht]{maniaSimpleRandomSearch2018}
Horia Mania, Aurelia Guy, and Benjamin Recht.
\newblock Simple random search provides a competitive approach to reinforcement
  learning.
\newblock \emph{arXiv:1803.07055}, 2018.

\bibitem[Mazoure et~al.(2020)Mazoure, Doan, Durand, Pineau, and
  Hjelm]{mazoureLeveragingExplorationOffpolicy2020}
Bogdan Mazoure, Thang Doan, Audrey Durand, Joelle Pineau, and R.~Devon Hjelm.
\newblock Leveraging exploration in off-policy algorithms via normalizing
  flows.
\newblock In \emph{Conference on {{Robot Learning}}}, pp.\  430--444, 2020.

\bibitem[Mnih et~al.(2015)Mnih, Kavukcuoglu, Silver, Rusu, Veness, Bellemare,
  Graves, Riedmiller, Fidjeland, Ostrovski, Petersen, Beattie, Sadik,
  Antonoglou, King, Kumaran, Wierstra, Legg, and
  Hassabis]{mnihHumanlevelControlDeep2015}
Volodymyr Mnih, Koray Kavukcuoglu, David Silver, Andrei~A. Rusu, Joel Veness,
  Marc~G. Bellemare, Alex Graves, Martin Riedmiller, Andreas~K. Fidjeland,
  Georg Ostrovski, Stig Petersen, Charles Beattie, Amir Sadik, Ioannis
  Antonoglou, Helen King, Dharshan Kumaran, Daan Wierstra, Shane Legg, and
  Demis Hassabis.
\newblock Human-level control through deep reinforcement learning.
\newblock \emph{Nature}, 518\penalty0 (7540):\penalty0 529--533, 2015.

\bibitem[Moore(1990)]{mooreEfficientMemorybasedLearning1990}
Andrew~William Moore.
\newblock {Efficient memory-based learning for robot control}.
\newblock Technical Report UCAM-CL-TR-209, University of Cambridge, Computer
  Laboratory, 1990.
\newblock URL \url{https://www.cl.cam.ac.uk/techreports/UCAM-CL-TR-209.pdf}.

\bibitem[Mutti et~al.(2020)Mutti, Pratissoli, and
  Restelli]{muttiPolicyGradientMethod2020}
Mirco Mutti, Lorenzo Pratissoli, and Marcello Restelli.
\newblock A {{Policy Gradient Method}} for {{Task-Agnostic Exploration}}.
\newblock \emph{arXiv:2007.04640}, 2020.

\bibitem[Nobakht \& Liu(2022)Nobakht and Liu]{nobakhtActionSpaceNoise2022}
Hesan Nobakht and Yong Liu.
\newblock Action space noise optimization as exploration in deterministic
  policy gradient for locomotion tasks.
\newblock \emph{Applied Intelligence}, 52\penalty0 (12):\penalty0 14218--14232,
  2022.

\bibitem[Noshad et~al.(2017)Noshad, Moon, Sekeh, and
  Hero]{noshadDirectEstimationInformation2017}
Morteza Noshad, Kevin~R. Moon, Salimeh~Yasaei Sekeh, and Alfred~O. Hero.
\newblock Direct estimation of information divergence using nearest neighbor
  ratios.
\newblock In \emph{{{IEEE International Symposium}} on {{Information Theory}}},
  pp.\  903--907, 2017.

\bibitem[Pinneri et~al.(2020)Pinneri, Sawant, Blaes, Achterhold, Stueckler,
  Rol{\'i}nek, and Martius]{pinneriSampleefficientCrossEntropyMethod2020}
Cristina Pinneri, Shambhuraj Sawant, Sebastian Blaes, Jan Achterhold, Joerg
  Stueckler, Michal Rol{\'i}nek, and Georg Martius.
\newblock Sample-efficient cross-entropy method for real-time planning.
\newblock In \emph{Conference on Robot Learning}, volume 155, pp.\  1049--1065,
  2020.

\bibitem[Plappert et~al.(2018)Plappert, Houthooft, Dhariwal, Sidor, Chen, Chen,
  Asfour, Abbeel, and Andrychowicz]{plappertParameterSpaceNoise2018}
Matthias Plappert, Rein Houthooft, Prafulla Dhariwal, Szymon Sidor, Richard~Y.
  Chen, Xi~Chen, Tamim Asfour, Pieter Abbeel, and Marcin Andrychowicz.
\newblock Parameter space noise for exploration.
\newblock In \emph{International Conference on Learning Representations}, 2018.

\bibitem[Pong et~al.(2020)Pong, Dalal, Lin, Nair, Bahl, and
  Levine]{pongSkewFitStateCoveringSelfSupervised2020}
Vitchyr Pong, Murtaza Dalal, Steven Lin, Ashvin Nair, Shikhar Bahl, and Sergey
  Levine.
\newblock Skew-fit: {{State-covering}} self-supervised reinforcement learning.
\newblock In \emph{International Conference on Machine Learning}, volume 119,
  pp.\  7783--7792, 2020.

\bibitem[Raffin(2020)]{rl-zoo3}
Antonin Raffin.
\newblock {{RL}} baselines3 zoo.
\newblock \emph{GitHub repository}, 2020.
\newblock URL \url{https://araffin.github.io/project/rl-baselines-zoo/}.

\bibitem[Raffin et~al.(2021{\natexlab{a}})Raffin, Hill, Gleave, Kanervisto,
  Ernestus, and Dormann]{stable-baselines3}
Antonin Raffin, Ashley Hill, Adam Gleave, Anssi Kanervisto, Maximilian
  Ernestus, and Noah Dormann.
\newblock Stable-baselines3: {{Reliable}} reinforcement learning
  implementations.
\newblock \emph{Journal of Machine Learning Research}, 22\penalty0
  (268):\penalty0 1--8, 2021{\natexlab{a}}.

\bibitem[Raffin et~al.(2021{\natexlab{b}})Raffin, Kober, and
  Stulp]{raffinSmoothExplorationRobotic2021}
Antonin Raffin, Jens Kober, and Freek Stulp.
\newblock Smooth exploration for robotic reinforcement learning.
\newblock In \emph{Conference on Robot Learning}, volume 164, pp.\  1634--1644,
  2021{\natexlab{b}}.

\bibitem[Ramsey et~al.(2011)Ramsey, Barrera, {Hachimine-Semprebom}, and
  Liu]{ramseyPairwiseComparisonsMeans2011}
Philip~H. Ramsey, Kyrstle Barrera, Pri {Hachimine-Semprebom}, and Chang-Chia
  Liu.
\newblock Pairwise comparisons of means under realistic nonnormality, unequal
  variances, outliers and equal sample sizes.
\newblock \emph{Journal of Statistical Computation and Simulation}, 81\penalty0
  (2):\penalty0 125--135, 2011.

\bibitem[Ramseyer \& Tcheng(1973)Ramseyer and
  Tcheng]{ramseyerRobustnessStudentizedRange1973}
Gary~C. Ramseyer and Tse-Kia Tcheng.
\newblock The {{Robustness}} of the {{Studentized Range Statistic}} to
  {{Violations}} of the {{Normality}} and {{Homogeneity}} of {{Variance
  Assumptions}}.
\newblock \emph{American Educational Research Journal}, 10\penalty0
  (3):\penalty0 235--240, 1973.

\bibitem[Rao et~al.(2020)Rao, Aljalbout, Sauer, and
  Haddadin]{raoHowMakeDeep2020}
Nirnai Rao, Elie Aljalbout, Axel Sauer, and Sami Haddadin.
\newblock How to {{Make Deep RL Work}} in {{Practice}}.
\newblock \emph{arXiv:2010.13083}, 2020.

\bibitem[Sauder \& DeMars(2019)Sauder and
  DeMars]{sauderUpdatedRecommendationMultiple2019}
Derek~C. Sauder and Christine~E. DeMars.
\newblock An {{Updated Recommendation}} for {{Multiple Comparisons}}.
\newblock \emph{Advances in Methods and Practices in Psychological Science},
  2\penalty0 (1):\penalty0 26--44, 2019.

\bibitem[Schulman et~al.(2015)Schulman, Levine, Moritz, Jordan, and
  Abbeel]{schulmanTrustRegionPolicy2015}
John Schulman, Sergey Levine, Philipp Moritz, Michael Jordan, and Pieter
  Abbeel.
\newblock Trust {{Region Policy Optimization}}.
\newblock In \emph{{{International Conference}} on {{Machine Learning}}}, pp.\
  1889--1897, 2015.

\bibitem[Schulman et~al.(2017)Schulman, Wolski, Dhariwal, Radford, and
  Klimov]{schulmanProximalPolicyOptimization2017}
John Schulman, Filip Wolski, Prafulla Dhariwal, Alec Radford, and Oleg Klimov.
\newblock Proximal {{Policy Optimization Algorithms}}.
\newblock \emph{arXiv:1707.06347}, 2017.

\bibitem[Seno \& Imai(2021)Seno and Imai]{senoD3rlpyOfflineDeep2021}
Takuma Seno and Michita Imai.
\newblock D3rlpy: {{An}} offline deep reinforcement library.
\newblock In \emph{{{NeurIPS}} 2021 Offline Reinforcement Learning Workshop},
  December 2021.

\bibitem[Seyde et~al.(2021)Seyde, Gilitschenski, Schwarting, Stellato,
  Riedmiller, Wulfmeier, and Rus]{seydeBangBangControlAll2021}
Tim Seyde, Igor Gilitschenski, Wilko Schwarting, Bartolomeo Stellato, Martin~A.
  Riedmiller, Markus Wulfmeier, and Daniela Rus.
\newblock Is bang-bang control all you need? {{Solving}} continuous control
  with bernoulli policies.
\newblock In \emph{Advances in Neural Information Processing Systems}, pp.\
  27209--27221, 2021.

\bibitem[Silver et~al.(2014)Silver, Lever, Heess, Degris, Wierstra, and
  Riedmiller]{silverDeterministicPolicyGradient2014}
David Silver, Guy Lever, Nicolas Heess, Thomas Degris, Daan Wierstra, and
  Martin Riedmiller.
\newblock Deterministic policy gradient algorithms.
\newblock In \emph{{{International Conference}} on {{Machine Learning}}}, pp.\
  387--395, 2014.

\bibitem[Sutton et~al.(1999)Sutton, McAllester, Singh, and
  Mansour]{suttonPolicyGradientMethods1999}
Richard~S Sutton, David McAllester, Satinder Singh, and Yishay Mansour.
\newblock Policy {{Gradient Methods}} for {{Reinforcement Learning}} with
  {{Function Approximation}}.
\newblock In \emph{Advances in {{Neural Information Processing Systems}}},
  1999.

\bibitem[Tang et~al.(2017)Tang, Houthooft, Foote, Stooke, Xi~Chen, Duan,
  Schulman, DeTurck, and Abbeel]{tangExplorationStudyCountBased2017}
Haoran Tang, Rein Houthooft, Davis Foote, Adam Stooke, OpenAI Xi~Chen, Yan
  Duan, John Schulman, Filip DeTurck, and Pieter Abbeel.
\newblock \#{{Exploration}}: {{A Study}} of {{Count-Based Exploration}} for
  {{Deep Reinforcement Learning}}.
\newblock In \emph{Advances in {{Neural Information Processing Systems}}}, pp.\
   2753--2762. 2017.

\bibitem[Tassa et~al.(2018)Tassa, Doron, Muldal, Erez, Li, Casas, Budden,
  Abdolmaleki, Merel, Lefrancq, Lillicrap, and
  Riedmiller]{tassaDeepMindControlSuite2018}
Yuval Tassa, Yotam Doron, Alistair Muldal, Tom Erez, Yazhe Li, Diego de~Las
  Casas, David Budden, Abbas Abdolmaleki, Josh Merel, Andrew Lefrancq, Timothy
  Lillicrap, and Martin Riedmiller.
\newblock {{DeepMind Control Suite}}.
\newblock \emph{arXiv:1801.00690}, 2018.

\bibitem[Todorov et~al.(2012)Todorov, Erez, and
  Tassa]{todorovMuJoCoPhysicsEngine2012}
Emanuel Todorov, Tom Erez, and Yuval Tassa.
\newblock {{MuJoCo}}: {{A}} physics engine for model-based control.
\newblock In \emph{{{IEEE}}/{{RSJ International Conference}} on {{Intelligent
  Robots}} and {{Systems}}}, pp.\  5026--5033, 2012.

\bibitem[Uhlenbeck \& Ornstein(1930)Uhlenbeck and
  Ornstein]{uhlenbeckTheoryBrownianMotion1930}
George~E. Uhlenbeck and Leonard~S. Ornstein.
\newblock On the theory of the {{Brownian}} motion.
\newblock \emph{Physical review}, 36\penalty0 (5):\penalty0 823, 1930.

\bibitem[Vallat(2018)]{vallatPingouinStatisticsPython2018}
Raphael Vallat.
\newblock Pingouin: Statistics in {{Python}}.
\newblock \emph{Journal of Open Source Software}, 3\penalty0 (31):\penalty0
  1026, 2018.

\bibitem[Ward et~al.(2019)Ward, Smofsky, and
  Bose]{wardImprovingExplorationSoftActorCritic2019}
Patrick~Nadeem Ward, Ariella Smofsky, and Avishek~Joey Bose.
\newblock Improving {{Exploration}} in {{Soft-Actor-Critic}} with {{Normalizing
  Flows Policies}}.
\newblock \emph{arXiv:1906.02771}, 2019.

\bibitem[Williams(1992)]{williamsSimpleStatisticalGradientfollowing1992}
Ronald~J. Williams.
\newblock Simple statistical gradient-following algorithms for connectionist
  reinforcement learning.
\newblock \emph{Machine learning}, 8\penalty0 (3-4):\penalty0 229--256, 1992.

\bibitem[Yang et~al.(2022)Yang, Tang, Bai, Liu, Hao, Meng, Liu, and
  Wang]{yangExplorationDeepReinforcement2022}
Tianpei Yang, Hongyao Tang, Chenjia Bai, Jinyi Liu, Jianye Hao, Zhaopeng Meng,
  Peng Liu, and Zhen Wang.
\newblock Exploration in {{Deep Reinforcement Learning}}: {{A Comprehensive
  Survey}}.
\newblock \emph{arXiv:2109.06668 [cs]}, July 2022.
\newblock \doi{10.48550/arXiv.2109.06668}.

\bibitem[Zhan et~al.(2019)Zhan, Aytemiz, and Smith]{zhanTakingScenicRoute2019}
Zeping Zhan, Batu Aytemiz, and Adam~M Smith.
\newblock Taking the scenic route: {{Automatic}} exploration for videogames.
\newblock In \emph{CEUR Workshop Proceedings}, pp.\  26--34, 2019.

\end{thebibliography}

\newpage
\appendix
\section*{Appendices}
\raggedbottom
\rhead{Appendix}
\counterwithin{table}{section}
\counterwithin{algorithm}{section}
\counterwithin{figure}{section}

\section{A motivating example \label{sec:motivating_example_details}}

The action is generated as $\tilde a_t \sim \pi_\theta(s_t)$,
$a_t = \tilde a_t + \varepsilon_{a_t}$, where $\varepsilon_{a_t}$
denotes the action noise.  We calibrate the noise scale to achieve
similar returns for both noise types.  To calibrate the action noise
scale, we assume a constant-zero-action policy upon which the action
noise is added and effectively use $a_t = \varepsilon_{a_t}$ as the
action sequence.  We find that a scale of about $0.6$ for Gaussian
action noise and a scale of about $0.5$ for Ornstein-Uhlenbeck noise
lead to a mean return of about $-30$.  This is shown in
\reftbl{tbl:mountaincar_motivation}.  A successful solution to the
Mountain-Car environment yields a positive return
$0 < \sum r_t < 100$.
We then use
these two noise configurations and perform learning with DDPG, SAC and
TD3.
The resulting learning curves are shown in
\reffig{fig:mountaincar_motivation_learning} and very clearly depict
the huge impact the noise configuration has: with similar returns of
the noise-only policies, we achieve substantially different learning
results, either leading to failure or success on the task.

To achieve a swing-up, the actions must not change direction too
rapidly but rather need to change direction with the right
frequency. Ornstein-Uhlenbeck noise is temporally correlated and thus helps
solving the environment successfully with a smaller scale $\sigma$. In
this environment, the algorithms tend to converge either to the
successful solution of the environment by swinging up, or to a passive
zero-action solution which incurs no penalty.

\begin{figure}[H]
\section{Boundary Artifacts}
  \floatbox[{\capbeside\thisfloatsetup{floatwidth=sidefil,capbesideposition={right,top},capbesidewidth=.7\linewidth}}]{figure}
  {\caption{\change{k-nearest-neighbor density estimators suffer from boundary
      artifacts when estimating densities with bounded support. The
      density around a query point is estimated by the volume required
      to include the $k$ nearest points. Top right, high density
      region: the volume to include $k$-points is smaller when the
      density is high. bottem left, low density: in lower densities a
      larger volume is required to include $k$ points. This also
      illustrates the boundary artifacts: when querying the density
      close to the support boundary, part of the query volume is
      outside the support. Thus the volume required to contain $k$
      points is over-estimated. This problem is amplified in higher
      dimensional spaces as the boundary artifacts occur as soon as
      any single dimension of the sphere protrudes outside the
      support.  }}\label{fig:boundary_artifacts}}
  {  \includegraphics[width=\linewidth]{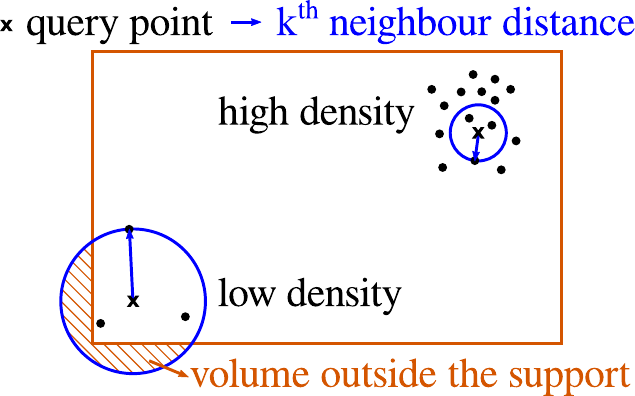}}
\end{figure}

\change{
\section{Action Noise in SAC}

SAC as defined by \cite{haarnojaSoftActorCriticAlgorithms2019} does
not use action noise for exploration. Instead, actions are sampled
from a stochastic Gaussian policy 
However, since SAC is an off-policy algorithm, additive action noise 
can additionally be used. The SAC algorithm uses
a target entropy parameter. The entropy coefficient of SAC is trained
such that the average entropy of the Gaussian policy matches this
target. In the implementation we use \citep{stable-baselines3}, the
entropy target can be automatically chosen based on the size of the
action space. In the Mountain-Car this amounts to a target entropy of
$1$. The entropy of a Gaussian is defined as
$ \mathcal{H}_{\mathcal{N}}(\sigma) =
\ln\left(\sigma\sqrt{2\,\pi\,e}\right) $. A $\sigma=1.7$ approximately
translates to an entropy target of $1.95$.

In SAC the value function $V$ contains an additional entropy bonus
term:
$ V(s_t) = \E_{a_t \sim \pi} \left[ Q(s_t, a_t) \right] + \alpha
\gH(\pi(\cdot|s_t)) $. This term is weighted by the entropy
coefficient $\alpha$.  Additionally, the SAC policy is defined as a
$\operatorname{soft max}$ operation over the Q function:
$ \pi_{\operatorname{softmax}}(a_t|s_t) = \frac {\exp(\frac{1}{\alpha}
  Q(s_t, a_t))} {Z(s_t)} $ where $Z$ is a normalizing term, chosen
s.t.  $\int \pi_{\text{softmax}}(a|s_t) \dif a = 1$.  Here, the
entropy coefficient $\alpha$ plays a double role, in both the entropy
bonus and the softness of the softmax operation. Thus, increasing the
scale of the Gaussian has a direct influence on the smoothness of the
softmax and can thus change the learning performance. Using action
noise is independent of the softmax and can be tuned
independently. Furthermore, action noise allows for the use of a
correlated noise process, which in the case of the Mountain-Car has a
large beneficial influence. This explains why using action noise can
be beneficial even for stochastic policies.
} 

\begin{table}[H]
  \change{
  \begin{tabular}{@{}lllllrrrr@{}}
\toprule
             &     &    &     &      &   P &     X &   R &     E \\
Environment & Algorithm & Type & Scale & Entropy Target &     &       &     &       \\
\midrule
\multirow{12}{*}{Mountain-Car} & \multirow{12}{*}{SAC} & \multirow{2}{*}{-} &     & 1.95 & -34 & -2.85 & -34 & -2.86 \\
             &     &    &     & auto &  -7 & -4.28 &  -7 & -4.29 \\
   \cmidrule[0pt]{3-9} 
 
             &     & \multirow{5}{*}{Gauss} & 0.1 & auto &  -5 & -4.15 &  -6 & -4.22 \\
             &     &    & 0.5 & auto &   3 & -2.94 & -18 & -3.93 \\
             &     &    & 0.9 & auto &  17 & -2.27 & -18 & -3.55 \\
             &     &    & 1.3 & auto &  23 & -2.06 & -21 & -3.38 \\
             &     &    & 1.7 & auto &  24 & -1.97 & -25 & -3.34 \\
   \cmidrule[0pt]{3-9} 
 
             &     & \multirow{5}{*}{OU} & 0.1 & auto &  -1 & -3.94 &  -3 & -4.11 \\
             &     &    & 0.5 & auto &  51 & -1.80 &  37 & -2.62 \\
             &     &    & 0.9 & auto &  68 & -1.49 &  53 & -2.22 \\
             &     &    & 1.3 & auto &  72 & -1.42 &  57 & -2.21 \\
             &     &    & 1.7 & auto &  73 & -1.39 &  57 & -2.14 \\
\bottomrule
\end{tabular}
   \caption{Comparison of SAC with Action noise against SAC relying on
    the stochastic policy for exploration. Increasing the entropy
    target increases the state space coverage.}
  }
\end{table}

\section{Deterministic SAC}
\begin{algorithm}[H]
  \caption{(Deterministic) Soft Actor-Critic}
  \label{alg:detsac}
  \begin{algorithmic}
    \State \mbox{Initialize parameter vectors $\psi$, ${\bar\psi}$, $\theta$, $\phi$.}
    \For{each iteration}
    \For{each environment step}
    \State $\mu_t, \sigma_t = f_\phi(s_t)$
    \State $\varepsilon_t \sim \mathcal{A}$ \Comment{$\mathcal{A} \ldots$ action noise process}
    \State {\color{ColDetsac} $a_t = \mu_t + \varepsilon_t$} \Comment{DetSAC}
    \vspace{0.5em}
    \State {\color{ColSac} $\pi_\phi(\cdot|s_t) = \mathcal{N}(\cdot | \mu_t, \sigma_t)$} \Comment{SAC}
    \State {\color{ColSac} $a'_t \sim \pi_\phi(\cdot|s_t)$}
    \State {\color{ColSac} $a_t = a'_t + \varepsilon_t $}
    \State $s_{t+1} \sim p(s_{t+1}| s_t, a_t)$
    \State $\mathcal{D} \leftarrow \mathcal{D} \cup \left\{(s_t, a_t, r(s_t, a_t), s_{t+1})\right\}$
    \EndFor
    \For{each gradient step}
    \State $\ldots$ \emph{original SAC update} \cite{haarnojaSoftActorCriticAlgorithms2019}
\EndFor
    \EndFor
  \end{algorithmic}
  \vfill
\end{algorithm}

\section{Benchmark Environments}
\begin{table}[H]
  \centering
\newcommand{\cs}{           & }  \begin{tabularx}{\columnwidth}{@{} m{3cm} @{\extracolsep{\fill}}m{2.8cm} m{.5cm} m{.5cm} m{8.cm} @{}}
    \toprule

    \sc{Environment}          & \sc{Illustration}                                                                                 & $\dim(\mathcal{O})$ & $\dim(\mathcal{A})$  & \sc{Reward}                                                                                                                                                                             \\
    \midrule
    Mountain-Car              & \includegraphics[width=2.5cm]{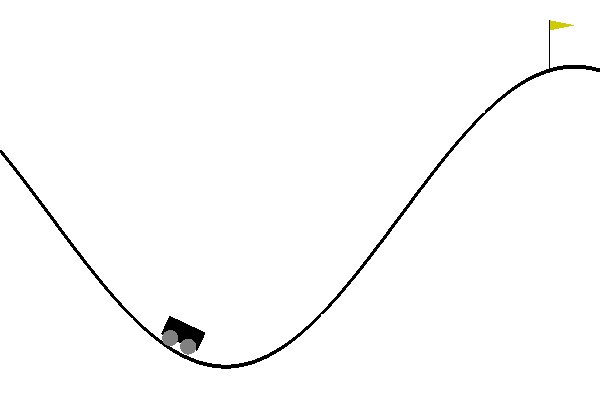}              & 2               & 1                & $\begin{aligned} \phantom{-} \cs \bm{1}(s_t, s_G)                            \cs -                     \cs |a_t|^2_2 \end{aligned}$                                                     \\
    Inverted-Pendulum-Swingup & \includegraphics[width=2.5cm]{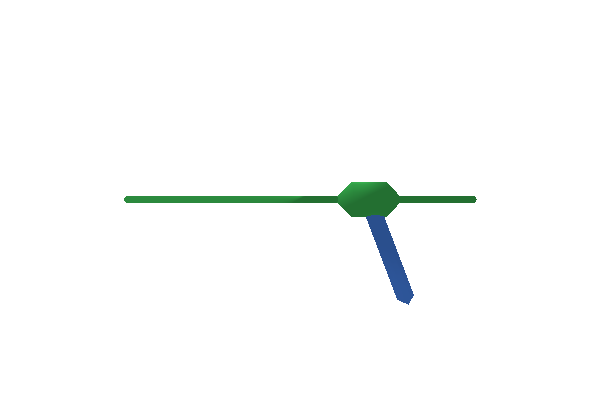} & 5               & 1                & $\begin{aligned} \phantom{-} \cs |\varphi(s_t) - \varphi_G|_1 \end{aligned}$                                                                                                            \\
    Reacher                   & \includegraphics[width=2.5cm]{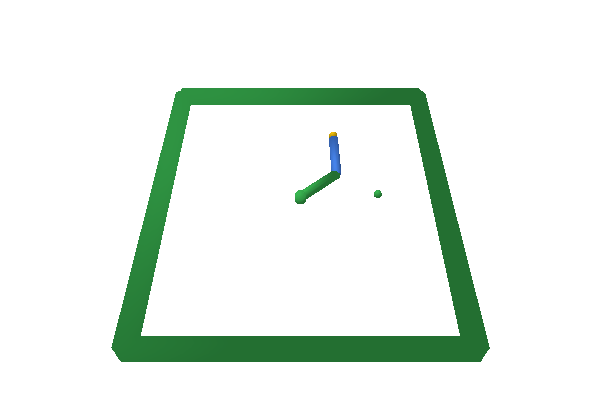}                 & 9               & 2                & $\begin{aligned}             \cs \nabla^{-}|s_t-s_G|_2                       \cs -  |\varphi(s_t)|_2^2 \cs ~-    \bm{1}(\varphi(s_t), \varphi_\text{limit}) \cs -  |a_t|_1\end{aligned}$\\
    Hopper                    & \includegraphics[width=2.5cm]{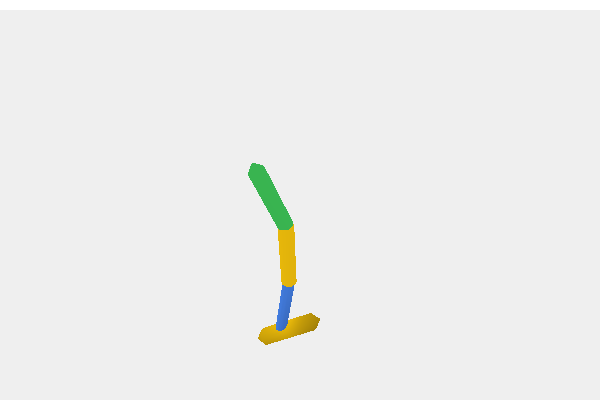}                  & 15              & 3                & $\begin{aligned}             \cs \nabla^{-}|s_t-s_G|_1                       \cs -  |\varphi(s_t)|_2^2 \cs ~-    \bm{1}(\varphi(s_t), \varphi_\text{limit}) \cs -  |a_t|_1\end{aligned}$\\
    Walker2D                  & \includegraphics[width=2.5cm]{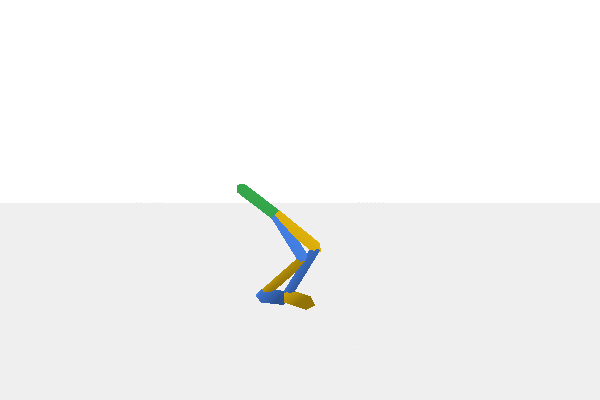}                & 22              & 6                & $\begin{aligned}             \cs \nabla^{-}|s_t-s_G|_1                       \cs -  |\varphi(s_t)|_2^2 \cs ~-    \bm{1}(\varphi(s_t), \varphi_\text{limit}) \cs -  |a_t|_1\end{aligned}$\\
    Half-Cheetah              & \includegraphics[width=2.5cm]{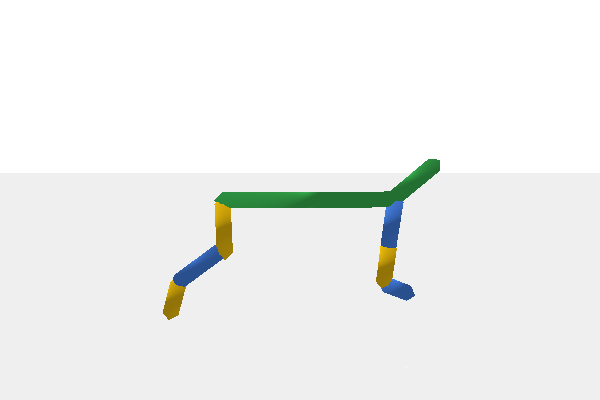}             & 26              & 6                & $\begin{aligned}             \cs \nabla^{-}|s_t-s_G|_1                       \cs -  |\varphi(s_t)|_2^2 \cs ~-    \bm{1}(\varphi(s_t), \varphi_\text{limit}) \cs -  |a_t|_1\end{aligned}$\\
    \bottomrule
  \end{tabularx}
\caption{Benchmarks environments used in our evaluation in increasing order of complexity.
    $|\mathcal{O}|$ denotes Observation space dimensions. $|\mathcal{A}|$ denotes Action space dimensions.
    Explanation of Reward components: $\bm{1}(b,c)$ indicator function (sparse
    reward or penalty) of $b$ w.r.t.\ to the set $c$; $|b|_n$ n-norm of
    $b$; $\varphi(s_t)$ angular component of state; $\nabla^{-} b$
    finite-difference reduction of $b$ between time-steps; $\varphi_\text{max}$ joint
    limit;  $s_G$ goal state; $|\varphi(s_t)|_2^2$ denotes an
    angular-power-penalty. Factors in the reward are omitted.
    Distances e.g.\ $|s_t-s_G|_n$ may refer to a subspace of the vector
    $s_t$.\label{tbl:envimgs} \refsec{dbg:environments}
  }
\end{table}

\section{Statistical Methods}

\begin{table}[H]
  \change{
    \begin{tabularx}{\textwidth}{@{}lllrrr@{}}
      \toprule
      Section                           & Where                                                     & Statistic           & $n / N$      & non-Normal         & $\neq$ Variance \\ \midrule
      \refsec{sec:q1_what_noise}        & \reffig{fig:WAN_noise_type_const}                         & Welch t-Test        & $2400/4800$  & CLT                & Robust          \\ & \reftbl{tbl:WAN_noise_type_envnames_const}                & Welch t-Test        & $400/800$    & CLT                & Robust          \\ \refsec{sec:q1_mann_whitney_u}    & \reftbl{tbl:WAN_noise_type_envnames_const_mann_whitney_u} & Mann-Whitney-U Test & $400/800$    & Robust             & Robust          \\ \refsec{sec:q3_should_schedule}   & \reftbl{tbl:scheduler_noise}                              & Levene's Test       & $800/2400$   & Robust             & -               \\ & \reftbl{tbl:scheduler_noise}                              & Games-Howell-Test   & $800/2400$   & $\alpha \cdot 0.2$ & Robust          \\ \refsec{sec:discuss_recommend}    & \reftbl{tbl:bestrank} (All) Scheduler                     & Games-Howell-Test   & $4800/14400$ & $\alpha \cdot 0.2$ & Robust          \\ \refsec{sec:discuss_recommend}    & \reftbl{tbl:bestrank} (All) Scale                         & Games-Howell-Test   & $2880/14400$ & $\alpha \cdot 0.2$ & Robust          \\ \refsec{sec:discuss_recommend}    & \reftbl{tbl:bestrank} (All) Type                          & Games-Howell-Test   & $3600/14400$ & $\alpha \cdot 0.2$ & Robust          \\ \refsec{sec:discuss_recommend}    & \reftbl{tbl:bestrank} (Env) Scheduler                     & Games-Howell-Test   & $800/2400$   & $\alpha \cdot 0.2$ & Robust          \\ \refsec{sec:discuss_recommend}    & \reftbl{tbl:bestrank} (Env) Scale                         & Games-Howell-Test   & $480/2400$   & $\alpha \cdot 0.2$ & Robust          \\ \refsec{sec:discuss_recommend}    & \reftbl{tbl:bestrank} (Env) Type                          & Games-Howell-Test   & $1200/2400$  & $\alpha \cdot 0.2$ & Robust          \\ \bottomrule
    \end{tabularx}
    \caption{\change{Summary of applied tests, per group sample size $n$ and
      cumulative size across groups $N$, see \refsec{sec:statmethods} about the
      $\alpha$ adjustment in the Games-Howell test. For large sample
      sizes the t-statistic approaches a normal distribution (CLT).
      Sample sizes of $30$ \citep{boneauEffectsViolationsAssumptions1960} are
      usually assumed to be large enough. \cite{lumleyImportanceNormalityAssumption2002}
      provide further evidence for the adequacy of our sample sizes.
      }}
  }
\end{table}
\subsection{Statistical Methods Details\label{sec:statmethods}}

We use statistical methods implemented in \cite{jonesSciPyOpenSource2001, vallatPingouinStatisticsPython2018} as well as our own implementations.

  \myparagraph{Welch t-test}: does not assume equal
  variance. Reporting two-tailed p-value. Significant for one-tailed
  when $\frac{p}{2} < \alpha$.
 \myparagraph{Games-Howell test} Performing multiple comparisons with a
 t-test increases the risk of Type I errors. To control for Type I
 errors, the Games-Howell test \change{\citep{gamesPairwiseMultipleComparison1976}, a multiple-comparison test applicable
 to cases with heterogeneity of variance, should be used \cite{sauderUpdatedRecommendationMultiple2019}.
 Sample sizes should be $n \geq 6$ in each group.

 The test statistic $t$ is distributed according to Tukey's
 studentized range $q$. \cite{gamesPairwiseMultipleComparison1976}
 describe that the test has been found to be robust to non-normality
 by \cite{ramseyerRobustnessStudentizedRange1973}, especially in the
 case of equal sample sizes. This holds in our
 case. \cite{ramseyPairwiseComparisonsMeans2011} have found the
 concurrent violation of homogeneity of variance and non-normality can
 increase type-I errors. Their results indicate that an error level of
 $\alpha=0.05$ can be achieved by applying a reduction of the
 significance level of $\alpha$ and find controlling for this error by
 reducing the significance level to $0.38\alpha$. Further evidence for
 reducing the significance threshold to $0.01$ in order to achieve
 error rates $<0.05$ is provided by
 \cite{ramseyerRobustnessStudentizedRange1973}.  }
\begin{align}
  t = & \frac{\bar{x}_i - \bar{x}_j}{\sigma} \\
\sigma = & \sqrt{{\left(\frac{s^2_i}{n_i} + \frac{s^2_j}{n_j}\right)}} \\
df = & \large \frac{\left(\frac{s^2_i}{n_i} +
\frac{s^2_j}{n_j}\right)^2}{\frac{\left(\frac{s_i^2}{n_i}\right)^2}{n_i - 1} +
         \frac{\left(\frac{s_j^2}{n_j}\right)^2}{n_j - 1}} \\
\end{align}

The $p$-value is then calculated for $k$ sample-groups as
\begin{align}
  q_{t\cdot \sqrt{2}, k, df}
\end{align}

 \myparagraph{ANOVA} We perform a \emph{balanced} N-way ANOVA,
 i.e. with N independent factors, each with multiple levels
 (categorical values). Since the study design is balanced this is
 equivalent to a type-I ANOVA in which the order of terms does not
 matter (because the design is balanced).

  \myparagraph{Eta squared $\bm{ \eta^2}$}

  The effect size eta squared $\eta^2$ denotes the relative variance
  explained by a factor to the total variance observed:
  $\eta^2 = \frac{SS_{C(x)}}{SS_\text{Total}}$

  \begin{table}[H]
  \begin{tabular}{lrrrrr}
    \toprule
    {}        & DF    & Sum of Squares & F       & PR(>F)  \\
    \midrule
    C(y)      & 9.0   & 4167.583       & 478.576 & 0       \\
    C(x)      & 9.0   & 91.118         & 10.463  & 1.7e-15 \\
    C(y):C(x) & 81.0  & 81.172         & 1.036   & 0.397   \\
    Residual  & 901.0 & 871.798        &         &         \\
    \midrule
    Total     &       & 5211.672       &         &         \\
    \bottomrule
    \caption{ANOVA example. The partial $\eta^2$ for a factor is calculated as the sum of squares, variance explained by that factor,
    divided by the sum of the variance explained plus the unexplained residual variance. }
    \label{tbl:anova_partial_eta_square}
  \end{tabular}
  \end{table}

  Effect sizes are interpreted as:
  \begin{align}
    \eta^2 \ge 0.01 & ~~~\textrm{small effect} \\
    \eta^2 \ge 0.06 & ~~~\textrm{medium effect} \\
    \eta^2 \ge 0.14 & ~~~\textrm{large effect} \\
  \end{align}

  \myparagraph{Levene's Test} assesses (un)equality of group variances.
  \begin{align}
    z_{ij} = & |y_{ij} - \tilde y _j| \\
    F = & \frac{N - p}{p - 1} \frac{\sum_{j=1}^p n_j(\tilde z_j - \tilde z)^2}{\sum_{j=1}^p \sum_{i=1}^{n_j} ( z_{ij} - \tilde z_j)^2} \\
    d_1 = & p - 1 \\
    d_2 = & N - p \\
    \tilde z_j = & \frac{1}{n_j} \sum_{i=1}^{n_j} z_{ij} \\
    \tilde z = & \frac{1}{N} \sum_{j=1}^p \sum_{i=1}^{n_j} z_{ij}
  \end{align}
  where $p$ is the number of groups, $n_j$ is the size of group $j$
  and $N$ is the total number of observations. $\tilde y_j$ is the
  median of group $j$, $z_{ij}$ denotes sample $i$ in group $j$. The
  $F$ statistic follows the F-distribution with degrees of freedom
  $d_1,d_2$.

  This variant of Levene's test, $\tilde y_j$ median instead of mean,
  is also called Brown-Forsythe test
  \cite{brownRobustTestsEquality1974} and is more robust to non-normal
  distributions.

 \myparagraph{Cohen-d effect size}: Cohen-d is illustrated in
 \reffig{fig:cohen_d} and measures the distance of the means of two
 sample groups normalized to the pooled variance:
\begin{figure}[H]
   \includegraphics[width=\columnwidth]{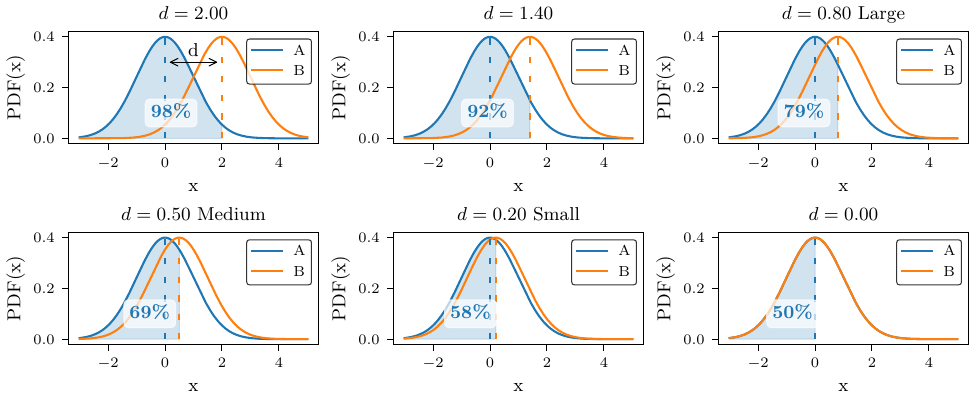}
   \caption{Illustration of Cohen-d effect size: the Cohen-d measures
     the standardized difference between the means of two groups,
     equivalent to a z-score. Effect sizes $d \ge 0.2$ are called
     small, $d \ge 0.5$ medium, $d \ge 0.8$ large effects. Under
     equal-variance Gaussian assumption this can be interpreted as
     $n$-percent of group A below the mean of group B. Illustrated as
     the shaded area.}
   \label{fig:cohen_d}
  \end{figure}
\begin{align}
     \textrm{Effect size} & = \frac{\textrm{[Mean Group A]} - \textrm{[Mean Group B]}}{\textrm{Pooled Std Deviation}} \\
   d & = \frac{\bar{x}_1 - \bar{x}_2} s \\
   s & = \sqrt{\frac{(n_1-1)s^2_1 + (n_2-1)s^2_2}{n_1+n_2 - 2}}
  \end{align}

\change{\section{(Q1) Which action noise type to use? -- Mann-Whitney-U Test \label{sec:q1_mann_whitney_u}}}

\begin{table}[H]
  \change{
  \begin{adjustbox}{angle=0,max width=\columnwidth,max height=\textheight}
    \begin{tabular}{@{}m{3cm}@{\hskip 0.3in}r@{~}c@{~}r@{}@{\hskip 0.3in}r@{~}c@{~}r@{}@{\hskip 0.3in}r@{~}c@{~}r@{}@{\hskip 0.3in}r@{~}c@{~}r@{}}
\toprule
              Environment &  P &                                 $p_P$ & $d_\text{P}$ &  R &                                 $p_R$ & $d_\text{R}$ &  X &                                 $p_X$ & $d_\text{X}$ &  E &                                 $p_E$ & $d_\text{E}$ \\
\midrule
             Half-Cheetah &  - &               {\footnotesize $0.08$ } &            - &  - &               {\footnotesize $0.86$ } &            - & OU &              {\footnotesize $0.005$ } &         0.21 &  - &               {\footnotesize $0.28$ } &            - \\
                   Hopper & OU &  {\footnotesize $\texttt{<}10^{-5}$ } &         0.27 &  G &  {\footnotesize $\texttt{<}10^{-4}$ } &         0.29 &  G & {\footnotesize $\texttt{<}10^{-10}$ } &         0.41 &  - &               {\footnotesize $0.69$ } &            - \\
Inverted-Pendulum-Swingup &  - &              {\footnotesize $0.040$ } &            - &  G & {\footnotesize $\texttt{<}10^{-40}$ } &         1.15 & OU & {\footnotesize $\texttt{<}10^{-47}$ } &         1.22 &  - &               {\footnotesize $0.32$ } &            - \\
             Mountain-Car & OU & {\footnotesize $\texttt{<}10^{-10}$ } &         0.47 & OU & {\footnotesize $\texttt{<}10^{-15}$ } &         0.66 & OU & {\footnotesize $\texttt{<}10^{-11}$ } &         0.34 & OU & {\footnotesize $\texttt{<}10^{-21}$ } &         0.71 \\
                  Reacher &  G & {\footnotesize $\texttt{<}10^{-32}$ } &         0.87 &  G & {\footnotesize $\texttt{<}10^{-21}$ } &         0.80 & OU & {\footnotesize $\texttt{<}10^{-35}$ } &         1.01 & OU & {\footnotesize $\texttt{<}10^{-26}$ } &         0.84 \\
                 Walker2D & OU &  {\footnotesize $\texttt{<}10^{-5}$ } &         0.15 &  G &              {\footnotesize $0.006$ } &         0.18 & OU & {\footnotesize $\texttt{<}10^{-13}$ } &         0.46 &  G &  {\footnotesize $\texttt{<}10^{-6}$ } &         0.08 \\
\bottomrule
\end{tabular}
   \end{adjustbox}
  \caption{\change{Comparison of noise types as in
    \reftbl{tbl:WAN_noise_type_envnames_const}, p-values of
    Mann-Whitney-U test are reported instead of the Welch-t-test.
    Similar tendencies are shown.}}
  \label{tbl:WAN_noise_type_envnames_const_mann_whitney_u}
  }
\end{table}

\section{Impact of Scheduler on Variance and Learned Performance}
\begin{table}[H]
  \begin{adjustbox}{angle=0,max width=\columnwidth,max height=\textheight}
    \begin{tabular}{@{}llccc@{\hskip 0.3in}ccc@{}}
\toprule
         &     & \multicolumn{3}{c}{$\text{var}(P)$} & \multicolumn{3}{c}{$P$} \\
         &     & {\footnotesize$<$ Constant} & {\footnotesize$<$ Linear} & {\footnotesize$<$ Logistic} & {\footnotesize$>$ Constant} & {\footnotesize$>$ Linear} & {\footnotesize$>$ Logistic} \\
Scheduler & Envname &                             &                           &                             &                             &                           &                             \\
\midrule
\multirow{6}{*}{Constant} & Half-Cheetah &                             &                        No &                          No &                             &                        No &                          No \\
         & Hopper &                             &                        No &                          No &                             &                        No &                          No \\
         & Inverted-Pendulum-Swingup &                             &                        No &                          No &                             &                        No &           Yes $p < 10^{-5}$ \\
         & Mountain-Car &                             &                        No &                          No &                             &         Yes $p < 10^{-6}$ &                          No \\
         & Reacher &                             &                        No &                          No &                             &                        No &                          No \\
         & Walker2D &                             &                        No &                          No &                             &                        No &                          No \\
  \\ 
 
\multirow{6}{*}{Linear} & Half-Cheetah &           Yes $p < 10^{-3}$ &                           &                          No &           Yes $p < 10^{-6}$ &                           &                          No \\
         & Hopper &           Yes $p < 10^{-6}$ &                           &                          No &           Yes $p < 10^{-6}$ &                           &                          No \\
         & Inverted-Pendulum-Swingup &                          No &                           &                          No &                          No &                           &                          No \\
         & Mountain-Car &                          No &                           &                          No &                          No &                           &                          No \\
         & Reacher &          Yes $p < 10^{-29}$ &                           &          Yes $p < 10^{-15}$ &           Yes $p < 10^{-6}$ &                           &           Yes $p < 10^{-6}$ \\
         & Walker2D &           Yes $p < 10^{-4}$ &                           &                          No &           Yes $p < 10^{-6}$ &                           &           Yes $p < 10^{-6}$ \\
  \\ 
 
\multirow{6}{*}{Logistic} & Half-Cheetah &           Yes $p < 10^{-4}$ &                        No &                             &           Yes $p < 10^{-6}$ &                        No &                             \\
         & Hopper &           Yes $p < 10^{-8}$ &                        No &                             &           Yes $p < 10^{-6}$ &                        No &                             \\
         & Inverted-Pendulum-Swingup &                          No &                        No &                             &                          No &                        No &                             \\
         & Mountain-Car &                          No &             Yes $p=0.002$ &                             &           Yes $p < 10^{-5}$ &         Yes $p < 10^{-6}$ &                             \\
         & Reacher &           Yes $p < 10^{-4}$ &                        No &                             &                          No &                        No &                             \\
         & Walker2D &           Yes $p < 10^{-6}$ &                        No &                             &           Yes $p < 10^{-4}$ &                        No &                             \\
  \\ 
 
Constant & Sum &                           0 &                         0 &                           0 &                           0 &                         1 &                           1 \\
Linear & Sum &                           4 &                         0 &                           1 &                           4 &                         0 &                           2 \\
Logistic & Sum &                           4 &                         1 &                           0 &                           4 &                         1 &                           0 \\
\bottomrule
\end{tabular}
   \end{adjustbox}
  \caption{In the majority of cases, using a scheduler \emph{reduces variance}
    \vvarP of the performance (evaluation returns), and \emph{improves} expected performance \vP.
The rows shows whether ``Scheduler''  is significantly better
    than the scheduler indicated in the columns \vvarP and
    \vP.
The evaluation returns \vP are standardized to control for the
    influence of algorithm, noise scale $\sigma$ and
    noise type. Levene's test is used to assess difference in variance
    \vvarP and a multiple-comparison Games-Howell test
    indicates superior performance \vP.  }
  \label{tbl:scheduler_noise_full}
\end{table}

\section{Performed experiments\label{sec:experiment_result_details}}

\change{This section lists the achieved final returns, calculated as
  the average of the evaluation returns of the last $5$ out of our
  $100$ training segments, for each noise setting, for the constant
  \reftbl{tbl:performed_experiments_constant}, linear
  \reftbl{tbl:performed_experiments_linear}, and logistic
  \reftbl{tbl:performed_experiments_constant} schedulers.

  Each noise configuration is repeated with $20$ random seeds.
  \reftbl{tbl:performed_experiments} lists the performance \vP for
  each noise configuration as the mean across the $20$ seeds.

  Choosing a fixed set of $20$ different seeds could introduce
  randomness artifacts, making one algorithm \emph{appear} to perform
  better than the other. To combat these biases, each run was
  performed from an independently, randomly drawn seed. The seeds are
  sampled using \texttt{os.urandom}, which provides a string of random
  bytes suitable for cryptographic use. This should be sufficient to
  ensure independence between seeds.
}

\begin{table}[H]
  \change{
    \begin{adjustbox}{angle=0,max width=.95\textwidth,max height=.8\textheight}
      \begin{tabular}{@{}lllrrrrrrrrrr@{}}
\toprule
         &          & {} & \multicolumn{10}{c}{Return} \\
         &          & Scale &    0.1 &   0.5 &   0.9 &   1.3 &   1.7 &  0.1 &  0.5 &  0.9 &  1.3 &  1.7 \\
         &          & Type &  Gauss & Gauss & Gauss & Gauss & Gauss &   OU &   OU &   OU &   OU &   OU \\
Scheduler & Environment & Algorithm &        &       &       &       &       &      &      &      &      &      \\
\midrule
\multirow{24}{*}{Constant} & \multirow{4}{*}{Half-Cheetah} & DDPG &    179 &   269 &   392 &   417 &   341 &  165 &  214 &  279 &  293 &  289 \\
         &          & DetSAC &   1719 &  1741 &   857 &   587 &   590 & 1619 & 1578 & 1156 &  787 &  731 \\
         &          & SAC &   1702 &  1856 &   531 &   389 &   292 & 1906 & 1550 & 1139 &  766 &  680 \\
         &          & TD3 &   1891 &  1651 &  1158 &   856 &   701 & 1582 & 1470 &  975 &  940 &  722 \\
   \cmidrule[0pt]{2-13} 
 
         & \multirow{4}{*}{Hopper} & DDPG &   1113 &   541 &   348 &   344 &   276 & 1131 &  808 &  666 &  507 &  461 \\
         &          & DetSAC &   2170 &  1101 &   837 &   572 &   542 & 2159 & 1602 & 1345 &  966 &  836 \\
         &          & SAC &   2291 &   979 &   808 &   717 &   604 & 2298 & 1599 & 1330 &  992 &  875 \\
         &          & TD3 &   2294 &  1833 &  1525 &  1370 &  1186 & 2258 & 1873 & 1539 & 1213 & 1164 \\
   \cmidrule[0pt]{2-13} 
 
         & \multirow{4}{*}{Inverted-Pendulum-Swingup} & DDPG &    819 &   819 &   816 &   827 &   818 &  822 &  859 &  879 &  877 &  878 \\
         &          & DetSAC &    838 &   884 &   880 &   886 &   887 &  842 &  887 &  886 &  888 &  888 \\
         &          & SAC &    881 &   887 &   883 &   888 &   884 &  886 &  888 &  889 &  887 &  888 \\
         &          & TD3 &    868 &   883 &   881 &   882 &   883 &  873 &  879 &  882 &  884 &  886 \\
   \cmidrule[0pt]{2-13} 
 
         & \multirow{4}{*}{Mountain-Car} & DDPG &     -0 &    52 &    84 &    67 &    74 &   -0 &   94 &   84 &   26 &   56 \\
         &          & DetSAC &      5 &    14 &    37 &    47 &    78 &   13 &   85 &   94 &   94 &   94 \\
         &          & SAC &      4 &    23 &    42 &    60 &    52 &    9 &   92 &   94 &   94 &   94 \\
         &          & TD3 &      5 &    94 &    65 &    74 &    74 &   -0 &   94 &   74 &   74 &   94 \\
   \cmidrule[0pt]{2-13} 
 
         & \multirow{4}{*}{Reacher} & DDPG &     17 &    18 &    18 &    18 &    17 &   16 &   15 &   14 &   13 &   13 \\
         &          & DetSAC &     19 &    19 &    19 &    19 &    19 &   18 &   18 &   17 &   16 &   15 \\
         &          & SAC &     18 &    18 &    17 &    16 &    16 &   18 &   17 &   16 &   15 &   13 \\
         &          & TD3 &     17 &    16 &    15 &    15 &    14 &   17 &   16 &   15 &   15 &   14 \\
   \cmidrule[0pt]{2-13} 
 
         & \multirow{4}{*}{Walker2D} & DDPG &    787 &   519 &   391 &   263 &   321 &  849 &  759 &  443 &  471 &  465 \\
         &          & DetSAC &   1814 &  1211 &   424 &   305 &   311 & 1616 & 1139 &  740 &  661 &  646 \\
         &          & SAC &   1883 &  1233 &   448 &   392 &   383 & 1710 & 1068 &  712 &  691 &  641 \\
         &          & TD3 &   2001 &  1878 &  1618 &  1321 &  1228 & 1821 & 1870 & 1596 & 1353 & 1230 \\
\bottomrule
\end{tabular}
     \end{adjustbox}
    \caption{This table shows the final evaluation return (mean over
      last five percent of training) for each configuration under the Constant regime. The mean
      across all $20$ runs is reported. \label{tbl:performed_experiments_constant} }
  }
\end{table}

\begin{table}[H]
  \change{
    \begin{adjustbox}{angle=0,max width=.95\textwidth,max height=.8\textheight}
      \begin{tabular}{@{}lllrrrrrrrrrr@{}}
\toprule
       &          & {} & \multicolumn{10}{c}{Return} \\
       &          & Scale &    0.1 &   0.5 &   0.9 &   1.3 &   1.7 &  0.1 &  0.5 &  0.9 &  1.3 &  1.7 \\
       &          & Type &  Gauss & Gauss & Gauss & Gauss & Gauss &   OU &   OU &   OU &   OU &   OU \\
Scheduler & Environment & Algorithm &        &       &       &       &       &      &      &      &      &      \\
\midrule
\multirow{24}{*}{Linear} & \multirow{4}{*}{Half-Cheetah} & DDPG &    158 &   155 &   154 &   150 &   142 &  176 &  142 &  138 &  127 &  124 \\
       &          & DetSAC &   1826 &  2538 &  1131 &   888 &   924 & 1547 & 1989 & 2062 & 1564 & 1657 \\
       &          & SAC &   1991 &  2471 &  1331 &   791 &   860 & 2249 & 2217 & 2062 & 1784 & 1735 \\
       &          & TD3 &   1957 &  2366 &  2034 &  1658 &  1731 & 1976 & 2069 & 2168 & 1881 & 1339 \\
   \cmidrule[0pt]{2-13} 
 
       & \multirow{4}{*}{Hopper} & DDPG &   1039 &  1071 &   883 &   864 &   790 &  965 &  937 &  999 &  964 &  871 \\
       &          & DetSAC &   2276 &  1923 &  1634 &  1540 &  1318 & 2308 & 1913 & 1702 & 1484 & 1297 \\
       &          & SAC &   2288 &  1840 &   905 &   814 &   789 & 2208 & 1899 & 1697 & 1579 & 1341 \\
       &          & TD3 &   2057 &  2107 &  1688 &  1581 &  1610 & 2099 & 2051 & 1943 & 1818 & 1633 \\
   \cmidrule[0pt]{2-13} 
 
       & \multirow{4}{*}{Inverted-Pendulum-Swingup} & DDPG &    836 &   830 &   832 &   838 &   835 &  823 &  817 &  829 &  830 &  750 \\
       &          & DetSAC &    867 &   882 &   885 &   878 &   882 &  778 &  887 &  887 &  888 &  888 \\
       &          & SAC &    884 &   883 &   879 &   881 &   888 &  879 &  888 &  886 &  888 &  887 \\
       &          & TD3 &    866 &   857 &   836 &   845 &   830 &  850 &  859 &  866 &  868 &  857 \\
   \cmidrule[0pt]{2-13} 
 
       & \multirow{4}{*}{Mountain-Car} & DDPG &     -0 &    56 &    46 &    31 &    65 &   -0 &   67 &   55 &   26 &   45 \\
       &          & DetSAC &     13 &    -0 &     5 &    28 &    38 &    5 &   33 &   95 &   95 &   95 \\
       &          & SAC &     -1 &     4 &     9 &    37 &    18 &    4 &   71 &   90 &   94 &   95 \\
       &          & TD3 &     -0 &    -0 &    37 &    56 &    84 &   -0 &   79 &   74 &   74 &   74 \\
   \cmidrule[0pt]{2-13} 
 
       & \multirow{4}{*}{Reacher} & DDPG &     16 &    16 &    16 &    16 &    16 &   16 &   16 &   16 &   16 &   15 \\
       &          & DetSAC &     18 &    19 &    20 &    19 &    20 &   19 &   18 &   17 &   17 &   17 \\
       &          & SAC &     18 &    18 &    18 &    18 &    18 &   18 &   18 &   17 &   17 &   17 \\
       &          & TD3 &     17 &    18 &    19 &    19 &    19 &   17 &   17 &   17 &   16 &   17 \\
   \cmidrule[0pt]{2-13} 
 
       & \multirow{4}{*}{Walker2D} & DDPG &    791 &   755 &   635 &   476 &   360 &  820 &  839 &  661 &  642 &  460 \\
       &          & DetSAC &   1602 &  1348 &  1048 &   970 &   894 & 1656 & 1276 & 1167 & 1014 &  876 \\
       &          & SAC &   1678 &  1630 &   845 &   637 &   652 & 1859 & 1334 & 1164 &  945 & 1028 \\
       &          & TD3 &   1780 &  1660 &  1584 &  1254 &  1238 & 1820 & 1766 & 1765 & 1647 & 1622 \\
\bottomrule
\end{tabular}
     \end{adjustbox}
    \caption{This table shows the final evaluation return (mean over
      last five percent of training) for each configuration under the Linear scheduler regime. The mean
      across all $20$ runs is reported. \label{tbl:performed_experiments_linear} }
  }
\end{table}

\begin{table}[H]
  \change{
    \begin{adjustbox}{angle=0,max width=.95\textwidth,max height=.8\textheight}
      \begin{tabular}{@{}lllrrrrrrrrrr@{}}
\toprule
         &          & {} & \multicolumn{10}{c}{Return} \\
         &          & Scale &    0.1 &   0.5 &   0.9 &   1.3 &   1.7 &  0.1 &  0.5 &  0.9 &  1.3 &  1.7 \\
         &          & Type &  Gauss & Gauss & Gauss & Gauss & Gauss &   OU &   OU &   OU &   OU &   OU \\
Scheduler & Environment & Algorithm &        &       &       &       &       &      &      &      &      &      \\
\midrule
\multirow{24}{*}{Logistic} & \multirow{4}{*}{Half-Cheetah} & DDPG &    158 &   156 &   143 &   136 &   124 &  180 &  153 &  129 &  128 &  108 \\
         &          & DetSAC &   1639 &  2168 &  1923 &  1559 &  1208 & 1680 & 2100 & 1921 & 1719 & 1235 \\
         &          & SAC &   1995 &  2199 &  1984 &  1746 &  1413 & 2158 & 2217 & 1927 & 1613 & 1511 \\
         &          & TD3 &   1694 &  2329 &  1864 &  1373 &  1363 & 2030 & 2047 & 1954 & 2002 & 1682 \\
   \cmidrule[0pt]{2-13} 
 
         & \multirow{4}{*}{Hopper} & DDPG &    902 &  1181 &  1103 &  1246 &  1147 & 1104 & 1057 & 1059 & 1207 & 1153 \\
         &          & DetSAC &   2237 &  1824 &  1691 &  1533 &  1338 & 2280 & 1822 & 1622 & 1482 & 1314 \\
         &          & SAC &   2130 &  1888 &  1654 &  1485 &  1385 & 2239 & 1855 & 1704 & 1469 & 1257 \\
         &          & TD3 &   2132 &  1922 &  1642 &  1796 &  1736 & 2037 & 2000 & 1786 & 1759 & 1694 \\
   \cmidrule[0pt]{2-13} 
 
         & \multirow{4}{*}{Inverted-Pendulum-Swingup} & DDPG &    837 &   832 &   841 &   821 &   676 &  756 &  842 &  832 &  747 &  845 \\
         &          & DetSAC &    839 &   887 &   887 &   888 &   889 &  792 &  886 &  888 &  889 &  889 \\
         &          & SAC &    873 &   888 &   888 &   889 &   889 &  876 &  889 &  888 &  888 &  888 \\
         &          & TD3 &    860 &   851 &   847 &   846 &   851 &  854 &  852 &  869 &  875 &  871 \\
   \cmidrule[0pt]{2-13} 
 
         & \multirow{4}{*}{Mountain-Car} & DDPG &     -0 &    70 &    74 &    36 &    45 &   -0 &   74 &   74 &   74 &   45 \\
         &          & DetSAC &      4 &    76 &    94 &    95 &    95 &    5 &   52 &   94 &   95 &   95 \\
         &          & SAC &      9 &    85 &    94 &    95 &    95 &    4 &   85 &   95 &   95 &   95 \\
         &          & TD3 &     -0 &     5 &    52 &    94 &    94 &   -0 &   94 &   79 &   84 &   84 \\
   \cmidrule[0pt]{2-13} 
 
         & \multirow{4}{*}{Reacher} & DDPG &     16 &    16 &    15 &    15 &    15 &   16 &   16 &   15 &   15 &   15 \\
         &          & DetSAC &     18 &    17 &    17 &    17 &    17 &   19 &   17 &   17 &   17 &   17 \\
         &          & SAC &     18 &    17 &    17 &    17 &    16 &   18 &   18 &   17 &   17 &   17 \\
         &          & TD3 &     17 &    19 &    19 &    19 &    19 &   17 &   17 &   16 &   16 &   16 \\
   \cmidrule[0pt]{2-13} 
 
         & \multirow{4}{*}{Walker2D} & DDPG &    905 &   792 &   659 &   426 &   341 &  792 &  885 &  676 &  457 &  324 \\
         &          & DetSAC &   1701 &  1371 &   986 &   891 &   789 & 1526 & 1201 &  923 &  914 &  767 \\
         &          & SAC &   1642 &  1274 &  1018 &   882 &   837 & 1676 & 1413 & 1148 &  850 &  805 \\
         &          & TD3 &   1759 &  1622 &  1294 &  1196 &   879 & 1842 & 1693 & 1611 & 1607 & 1630 \\
\bottomrule
\end{tabular}
     \end{adjustbox}
    \caption{This table shows the final evaluation return (mean over
      last five percent of training) for each configuration under the Logistic scheduler regime. The mean
      across all $20$ runs is reported. \label{tbl:performed_experiments_logistic} }
  }
\end{table}

\begin{table}[H]
  \change{
\begin{adjustbox}{angle=0,totalheight=.95\textheight,max width=1.\textwidth}
      \begin{tabular}{@{}lllrrrrrrrrrr@{}}
\toprule
         &          & {} & \multicolumn{10}{c}{P} \\
         &          & Scale &   0.1 &   0.5 &   0.9 &   1.3 &   1.7 &  0.1 &  0.5 &  0.9 &  1.3 &  1.7 \\
         &          & Type & Gauss & Gauss & Gauss & Gauss & Gauss &   OU &   OU &   OU &   OU &   OU \\
Scheduler & Environment & Algorithm &       &       &       &       &       &      &      &      &      &      \\
\midrule
\multirow{24}{*}{Constant} & \multirow{4}{*}{Half-Cheetah} & DDPG &   192 &   218 &   343 &   322 &   259 &  174 &  204 &  254 &  238 &  249 \\
         &          & DetSAC &  1148 &  1272 &   743 &   594 &   577 & 1026 & 1070 &  869 &  690 &  651 \\
         &          & SAC &  1109 &  1413 &   567 &   437 &   354 & 1158 & 1102 &  848 &  676 &  621 \\
         &          & TD3 &  1284 &  1179 &   885 &   680 &   603 & 1053 & 1052 &  760 &  770 &  626 \\
   \cmidrule[0pt]{2-13} 
 
         & \multirow{4}{*}{Hopper} & DDPG &   950 &   498 &   343 &   288 &   222 &  946 &  717 &  572 &  407 &  321 \\
         &          & DetSAC &  1957 &  1100 &   814 &   671 &   632 & 1927 & 1458 & 1205 &  903 &  817 \\
         &          & SAC &  1976 &  1108 &   813 &   746 &   690 & 2044 & 1464 & 1215 &  906 &  813 \\
         &          & TD3 &  1911 &  1547 &  1199 &  1005 &   804 & 1961 & 1575 & 1178 &  963 &  776 \\
   \cmidrule[0pt]{2-13} 
 
         & \multirow{4}{*}{Inverted-Pendulum-Swingup} & DDPG &   738 &   751 &   746 &   748 &   743 &  738 &  760 &  755 &  753 &  760 \\
         &          & DetSAC &   793 &   857 &   855 &   858 &   864 &  815 &  843 &  840 &  842 &  842 \\
         &          & SAC &   841 &   860 &   856 &   863 &   860 &  827 &  850 &  850 &  846 &  847 \\
         &          & TD3 &   703 &   763 &   748 &   761 &   749 &  705 &  758 &  749 &  755 &  751 \\
   \cmidrule[0pt]{2-13} 
 
         & \multirow{4}{*}{Mountain-Car} & DDPG &    -0 &    44 &    62 &    59 &    65 &   -0 &   87 &   80 &   23 &   53 \\
         &          & DetSAC &     1 &     6 &    15 &    23 &    41 &    4 &   52 &   73 &   78 &   80 \\
         &          & SAC &    -5 &     3 &    17 &    23 &    24 &   -1 &   51 &   68 &   72 &   73 \\
         &          & TD3 &     2 &    82 &    58 &    70 &    70 &   -0 &   77 &   69 &   67 &   87 \\
   \cmidrule[0pt]{2-13} 
 
         & \multirow{4}{*}{Reacher} & DDPG &    14 &    15 &    15 &    15 &    15 &   14 &   12 &   11 &   10 &    9 \\
         &          & DetSAC &    17 &    18 &    17 &    17 &    18 &   16 &   15 &   13 &   11 &   10 \\
         &          & SAC &    16 &    16 &    16 &    14 &    13 &   16 &   15 &   13 &   11 &    9 \\
         &          & TD3 &    15 &    12 &    11 &    10 &     9 &   14 &   12 &   11 &   10 &    9 \\
   \cmidrule[0pt]{2-13} 
 
         & \multirow{4}{*}{Walker2D} & DDPG &   484 &   367 &   280 &   243 &   239 &  498 &  455 &  327 &  336 &  346 \\
         &          & DetSAC &  1324 &  1034 &   448 &   301 &   293 & 1175 &  846 &  674 &  625 &  594 \\
         &          & SAC &  1413 &  1103 &   406 &   369 &   367 & 1290 &  824 &  658 &  600 &  588 \\
         &          & TD3 &  1504 &  1419 &  1098 &   842 &   730 & 1491 & 1440 & 1137 &  885 &  717 \\
   \cmidrule[0pt]{1-13} 
 
   \cmidrule[0pt]{2-13} 
 
\multirow{24}{*}{Linear} & \multirow{4}{*}{Half-Cheetah} & DDPG &   172 &   178 &   187 &   177 &   163 &  187 &  162 &  165 &  169 &  167 \\
         &          & DetSAC &  1230 &  1923 &   913 &   743 &   707 & 1030 & 1230 & 1186 &  907 &  905 \\
         &          & SAC &  1248 &  1840 &   966 &   654 &   627 & 1435 & 1346 & 1260 &  958 &  899 \\
         &          & TD3 &  1300 &  1727 &  1360 &  1099 &  1076 & 1292 & 1288 & 1243 & 1086 &  763 \\
   \cmidrule[0pt]{2-13} 
 
         & \multirow{4}{*}{Hopper} & DDPG &   896 &   769 &   607 &   527 &   498 &  877 &  774 &  696 &  594 &  546 \\
         &          & DetSAC &  2076 &  1551 &  1302 &  1174 &   977 & 2035 & 1667 & 1356 & 1131 &  969 \\
         &          & SAC &  1966 &  1554 &   783 &   713 &   701 & 1977 & 1601 & 1346 & 1117 &  999 \\
         &          & TD3 &  1772 &  1602 &   996 &   854 &   920 & 1854 & 1688 & 1452 & 1315 & 1177 \\
   \cmidrule[0pt]{2-13} 
 
         & \multirow{4}{*}{Inverted-Pendulum-Swingup} & DDPG &   713 &   749 &   757 &   756 &   748 &  729 &  753 &  755 &  751 &  671 \\
         &          & DetSAC &   818 &   861 &   857 &   857 &   859 &  774 &  844 &  838 &  840 &  840 \\
         &          & SAC &   850 &   860 &   857 &   858 &   862 &  835 &  847 &  849 &  848 &  848 \\
         &          & TD3 &   696 &   748 &   746 &   745 &   740 &  720 &  746 &  743 &  755 &  747 \\
   \cmidrule[0pt]{2-13} 
 
         & \multirow{4}{*}{Mountain-Car} & DDPG &    -0 &    46 &    39 &    28 &    60 &   -0 &   64 &   52 &   24 &   40 \\
         &          & DetSAC &     2 &    -0 &     2 &    16 &    23 &    1 &   23 &   75 &   79 &   77 \\
         &          & SAC &    -7 &    -3 &    -1 &    16 &     7 &   -3 &   45 &   59 &   68 &   73 \\
         &          & TD3 &    -0 &    -0 &    33 &    46 &    73 &   -0 &   69 &   68 &   69 &   68 \\
   \cmidrule[0pt]{2-13} 
 
         & \multirow{4}{*}{Reacher} & DDPG &    14 &    14 &    14 &    13 &    13 &   14 &   12 &   12 &   12 &   11 \\
         &          & DetSAC &    17 &    18 &    18 &    18 &    19 &   16 &   15 &   14 &   13 &   11 \\
         &          & SAC &    16 &    17 &    17 &    17 &    17 &   16 &   15 &   14 &   13 &   11 \\
         &          & TD3 &    15 &    17 &    17 &    17 &    17 &   14 &   12 &   12 &   11 &   11 \\
   \cmidrule[0pt]{2-13} 
 
         & \multirow{4}{*}{Walker2D} & DDPG &   484 &   467 &   372 &   293 &   256 &  509 &  495 &  371 &  331 &  272 \\
         &          & DetSAC &  1233 &   976 &   800 &   742 &   705 & 1206 &  973 &  860 &  757 &  689 \\
         &          & SAC &  1235 &  1291 &   622 &   510 &   491 & 1391 &  972 &  852 &  713 &  742 \\
         &          & TD3 &  1477 &  1384 &   918 &   665 &   612 & 1457 & 1397 & 1273 & 1176 & 1072 \\
   \cmidrule[0pt]{1-13} 
 
   \cmidrule[0pt]{2-13} 
 
\multirow{24}{*}{Logistic} & \multirow{4}{*}{Half-Cheetah} & DDPG &   175 &   168 &   166 &   166 &   162 &  182 &  177 &  160 &  167 &  165 \\
         &          & DetSAC &  1059 &  1358 &  1094 &   875 &   781 & 1065 & 1350 & 1122 &  967 &  768 \\
         &          & SAC &  1269 &  1433 &  1094 &   943 &   822 & 1329 & 1418 & 1115 &  912 &  842 \\
         &          & TD3 &  1224 &  1700 &  1292 &   943 &   931 & 1260 & 1234 & 1126 & 1109 &  915 \\
   \cmidrule[0pt]{2-13} 
 
         & \multirow{4}{*}{Hopper} & DDPG &   874 &   874 &   727 &   662 &   586 &  932 &  835 &  713 &  659 &  589 \\
         &          & DetSAC &  2064 &  1543 &  1361 &  1139 &   988 & 2027 & 1586 & 1268 & 1092 &  962 \\
         &          & SAC &  1911 &  1533 &  1306 &  1096 &  1000 & 2057 & 1589 & 1322 & 1073 &  941 \\
         &          & TD3 &  1827 &  1337 &   845 &   932 &   939 & 1803 & 1650 & 1358 & 1220 & 1104 \\
   \cmidrule[0pt]{2-13} 
 
         & \multirow{4}{*}{Inverted-Pendulum-Swingup} & DDPG &   726 &   756 &   750 &   757 &   589 &  656 &  748 &  759 &  676 &  739 \\
         &          & DetSAC &   794 &   841 &   838 &   842 &   843 &  777 &  846 &  845 &  843 &  840 \\
         &          & SAC &   843 &   850 &   849 &   849 &   845 &  838 &  856 &  852 &  848 &  840 \\
         &          & TD3 &   717 &   753 &   756 &   753 &   750 &  712 &  752 &  751 &  746 &  740 \\
   \cmidrule[0pt]{2-13} 
 
         & \multirow{4}{*}{Mountain-Car} & DDPG &    -0 &    66 &    70 &    32 &    42 &   -0 &   66 &   72 &   70 &   42 \\
         &          & DetSAC &     0 &    51 &    73 &    76 &    79 &    2 &   37 &   75 &   79 &   79 \\
         &          & SAC &    -0 &    52 &    62 &    68 &    71 &   -4 &   48 &   72 &   70 &   72 \\
         &          & TD3 &    -0 &     3 &    39 &    71 &    81 &   -0 &   81 &   71 &   80 &   80 \\
   \cmidrule[0pt]{2-13} 
 
         & \multirow{4}{*}{Reacher} & DDPG &    14 &    12 &    11 &    11 &    10 &   14 &   12 &   12 &   11 &   10 \\
         &          & DetSAC &    16 &    14 &    13 &    12 &    11 &   16 &   15 &   13 &   11 &   11 \\
         &          & SAC &    16 &    15 &    14 &    12 &    11 &   16 &   15 &   14 &   11 &   10 \\
         &          & TD3 &    15 &    17 &    17 &    17 &    17 &   15 &   13 &   12 &   11 &   11 \\
   \cmidrule[0pt]{2-13} 
 
         & \multirow{4}{*}{Walker2D} & DDPG &   508 &   468 &   352 &   269 &   237 &  498 &  462 &  350 &  267 &  242 \\
         &          & DetSAC &  1266 &   992 &   751 &   686 &   642 & 1146 &  922 &  732 &  699 &  650 \\
         &          & SAC &  1232 &   955 &   766 &   672 &   652 & 1237 & 1022 &  795 &  680 &  636 \\
         &          & TD3 &  1442 &  1313 &   672 &   608 &   476 & 1460 & 1326 & 1154 & 1000 &  982 \\
\bottomrule
\end{tabular}
     \end{adjustbox}
    \caption{\change{This table shows the performance P as the mean
        across the $20$ different seeds.}\label{tbl:performed_experiments} }
  }
\end{table}

\section{Hyperparameters}

\begin{table}[H]
  \caption{Hyperparameters for SAC, TD3 and DDPG are taken from
    \cite{rl-zoo3} or left at default values defined in
    \cite{stable-baselines3}.}
\begin{subtable}[h]{\textwidth}
  \begin{adjustbox}{angle=0,max width=.95\textwidth,max height=.9\textheight}
    \change{\begin{tabular}{@{}lp{0.16\textwidth}<{\raggedright}p{0.16\textwidth}<{\raggedright}p{0.16\textwidth}<{\raggedright}p{0.16\textwidth}<{\raggedright}p{0.16\textwidth}<{\raggedright}p{0.16\textwidth}<{\raggedright}@{}}
\toprule
Environment & Walker2D & Inverted-Pendulum-Swingup & Hopper & Mountain-Car & Half-Cheetah & Reacher \\
\midrule
env\_wrapper & \bfseries Time\allowbreak Feature\allowbreak Wrapper & \bfseries  & \bfseries Time\allowbreak Feature\allowbreak Wrapper & \bfseries  & \bfseries Time\allowbreak Feature\allowbreak Wrapper & \bfseries  \\
gamma & 0.99 & 0.99 & 0.99 & 0.99 & \bfseries 1 & 0.99 \\
buffer\_size & 1000000 & 1000000 & 1000000 & \bfseries 50000 & 1000000 & 1000000 \\
learning\_starts & \bfseries 1000 & \bfseries 1000 & \bfseries 1000 & \bfseries 0 & \bfseries 10000 & \bfseries 1000 \\
gradient\_steps & \bfseries -1 & \bfseries -1 & \bfseries -1 & 1 & \bfseries -1 & \bfseries -1 \\
train\_freq & \bfseries (1, 'episode') & \bfseries (1, 'episode') & \bfseries (1, 'episode') & 1 & \bfseries (1, 'episode') & \bfseries (1, 'episode') \\
learning\_rate & 0.0003 & 0.0003 & 0.0003 & 0.0003 & 0.0003 & 0.0003 \\
timesteps & \bfseries 2000000 & \bfseries 1000000 & \bfseries 2000000 & \bfseries 60000 & \bfseries 2000000 & \bfseries 1000000 \\
ID & \bfseries \footnotesize Walker2DBullet\allowbreak Env-v0 & \bfseries \footnotesize Inverted\allowbreak Pendulum\allowbreak Swingup\allowbreak PyBullet\allowbreak Env-v0 & \bfseries \footnotesize Hopper\allowbreak PyBullet\allowbreak Env-v0 & \bfseries \footnotesize MountainCar\allowbreak Continuous-v0 & \bfseries \footnotesize HalfCheetah\allowbreak PyBullet\allowbreak Env-v0 & \bfseries \footnotesize Reacher\allowbreak PyBullet\allowbreak Env-v0 \\
batch\_size & 256 & \bfseries 64 & 256 & \bfseries 64 & 256 & \bfseries 64 \\
ent\_coef & \bfseries 0 & \bfseries 0 & \bfseries 0 & auto & auto & \bfseries 0 \\
tau & 0.005 & 0.005 & 0.005 & 0.005 & \bfseries 0 & 0.005 \\
\bottomrule
\end{tabular}
 }
  \end{adjustbox}
  \caption{SAC/DetSAC Hyperparameters}
  \vspace{1em}
\end{subtable}

\begin{subtable}[h]{\textwidth}
  \begin{adjustbox}{angle=0,max width=0.95\textwidth,max height=.9\textheight}
    \change{\begin{tabular}{@{}lp{0.16\textwidth}<{\raggedright}p{0.16\textwidth}<{\raggedright}p{0.16\textwidth}<{\raggedright}p{0.16\textwidth}<{\raggedright}p{0.16\textwidth}<{\raggedright}p{0.16\textwidth}<{\raggedright}@{}}
\toprule
Environment & Walker2D & Inverted-Pendulum-Swingup & Hopper & Mountain-Car & Half-Cheetah & Reacher \\
\midrule
env\_wrapper & \bfseries Time\allowbreak Feature\allowbreak Wrapper & \bfseries Time\allowbreak Feature\allowbreak Wrapper & \bfseries Time\allowbreak Feature\allowbreak Wrapper & \bfseries  & \bfseries Time\allowbreak Feature\allowbreak Wrapper & \bfseries Time\allowbreak Feature\allowbreak Wrapper \\
gamma & \bfseries 1 & \bfseries 1 & \bfseries 1 & 0.99 & \bfseries 1 & \bfseries 1 \\
buffer\_size & \bfseries 200000 & \bfseries 200000 & \bfseries 200000 & 1000000 & \bfseries 200000 & \bfseries 200000 \\
learning\_starts & \bfseries 10000 & \bfseries 10000 & \bfseries 10000 & 100 & \bfseries 10000 & \bfseries 10000 \\
gradient\_steps & -1 & -1 & -1 & -1 & -1 & -1 \\
train\_freq & (1, 'episode') & (1, 'episode') & (1, 'episode') & (1, 'episode') & (1, 'episode') & (1, 'episode') \\
learning\_rate & 0.001 & 0.001 & 0.001 & 0.001 & 0.001 & 0.001 \\
policy\_kwargs & \bfseries \{'net\_arch': [400, 300]\} & \bfseries \{'net\_arch': [400, 300]\} & \bfseries \{'net\_arch': [400, 300]\} & None & \bfseries \{'net\_arch': [400, 300]\} & \bfseries \{'net\_arch': [400, 300]\} \\
timesteps & \bfseries 1000000 & \bfseries 300000 & \bfseries 1000000 & \bfseries 300000 & \bfseries 1000000 & \bfseries 300000 \\
ID & \bfseries \footnotesize Walker2DBullet\allowbreak Env-v0 & \bfseries \footnotesize Inverted\allowbreak Pendulum\allowbreak Swingup\allowbreak PyBullet\allowbreak Env-v0 & \bfseries \footnotesize Hopper\allowbreak PyBullet\allowbreak Env-v0 & \bfseries \footnotesize MountainCar\allowbreak Continuous-v0 & \bfseries \footnotesize HalfCheetah\allowbreak PyBullet\allowbreak Env-v0 & \bfseries \footnotesize Reacher\allowbreak PyBullet\allowbreak Env-v0 \\
\bottomrule
\end{tabular}
 }
  \end{adjustbox}
  \caption{TD3 Hyperparameters}
  \vspace{1em}
\end{subtable}
\begin{subtable}[h]{\textwidth}
  \begin{adjustbox}{angle=0,max width=0.95\textwidth,max height=.9\textheight}
    \change{\begin{tabular}{@{}lp{0.16\textwidth}<{\raggedright}p{0.16\textwidth}<{\raggedright}p{0.16\textwidth}<{\raggedright}p{0.16\textwidth}<{\raggedright}p{0.16\textwidth}<{\raggedright}p{0.16\textwidth}<{\raggedright}@{}}
\toprule
Environment & Walker2D & Inverted-Pendulum-Swingup & Hopper & Mountain-Car & Half-Cheetah & Reacher \\
\midrule
env\_wrapper & \bfseries Time\allowbreak Feature\allowbreak Wrapper & \bfseries Time\allowbreak Feature\allowbreak Wrapper & \bfseries Time\allowbreak Feature\allowbreak Wrapper & \bfseries  & \bfseries Time\allowbreak Feature\allowbreak Wrapper & \bfseries Time\allowbreak Feature\allowbreak Wrapper \\
gamma & \bfseries 1 & \bfseries 1 & \bfseries 1 & 0.99 & \bfseries 1 & \bfseries 1 \\
buffer\_size & 1000000 & \bfseries 200000 & 1000000 & 1000000 & \bfseries 200000 & \bfseries 200000 \\
learning\_starts & \bfseries 10000 & \bfseries 10000 & \bfseries 10000 & 100 & \bfseries 10000 & \bfseries 10000 \\
gradient\_steps & -1 & -1 & -1 & -1 & -1 & -1 \\
train\_freq & (1, 'episode') & (1, 'episode') & (1, 'episode') & (1, 'episode') & (1, 'episode') & \bfseries 1 \\
learning\_rate & \bfseries 0.0007 & 0.001 & \bfseries 0.0007 & 0.001 & 0.001 & 0.001 \\
policy\_kwargs & \bfseries \{'net\_arch': [400, 300]\} & \bfseries \{'net\_arch': [400, 300]\} & \bfseries \{'net\_arch': [400, 300]\} & None & \bfseries \{'net\_arch': [400, 300]\} & \bfseries \{'net\_arch': [400, 300]\} \\
timesteps & \bfseries 1000000 & \bfseries 300000 & \bfseries 1000000 & \bfseries 300000 & \bfseries 1000000 & \bfseries 300000 \\
ID & \bfseries \footnotesize Walker2DBullet\allowbreak Env-v0 & \bfseries \footnotesize Inverted\allowbreak Pendulum\allowbreak Swingup\allowbreak PyBullet\allowbreak Env-v0 & \bfseries \footnotesize Hopper\allowbreak PyBullet\allowbreak Env-v0 & \bfseries \footnotesize MountainCar\allowbreak Continuous-v0 & \bfseries \footnotesize HalfCheetah\allowbreak PyBullet\allowbreak Env-v0 & \bfseries \footnotesize Reacher\allowbreak PyBullet\allowbreak Env-v0 \\
batch\_size & \bfseries 256 & 100 & \bfseries 256 & 100 & 100 & 100 \\
\bottomrule
\end{tabular}
 }
  \end{adjustbox}
  \caption{DDPG Hyperparameters}
\end{subtable}
\end{table}

\section{Environment Limits}

\begin{table}[H]
  \begin{adjustbox}{angle=0,max width=1.0\textwidth,max height=.9\textheight}
    \begin{tabular}{rcccccc}
\toprule
Environment &               MountainCarContinuous-v0 &  InvertedPendulumSwingupPyBulletEnv-v0 &                  ReacherPyBulletEnv-v0 &                   HopperPyBulletEnv-v0 &                   Walker2DBulletEnv-v0 &              HalfCheetahPyBulletEnv-v0 \\
\midrule
$s^{(0)}$  &  $  \textrm{-}1.2000 \ldots    0.6000$ &  $  \textrm{-}1.0993 \ldots    1.0931$ &  $  \textrm{-}0.2700 \ldots    0.2700$ &  $  \textrm{-}1.2433 \ldots    0.8614$ &  $  \textrm{-}1.2316 \ldots    0.1270$ &  $  \textrm{-}0.6542 \ldots    0.5536$ \\
$s^{(1)}$  &  $  \textrm{-}0.0700 \ldots    0.0700$ &  $  \textrm{-}6.1276 \ldots    6.0216$ &  $  \textrm{-}0.2700 \ldots    0.2700$ &  $  \textrm{-}0.0000 \ldots    0.0000$ &  $  \textrm{-}0.0000 \ldots    0.0000$ &  $  \textrm{-}0.0000 \ldots    0.0000$ \\
$s^{(2)}$  &                                        &  $  \textrm{-}1.0000 \ldots    1.0000$ &  $  \textrm{-}0.4799 \ldots    0.4798$ &  $  \textrm{-}1.0000 \ldots    1.0000$ &  $  \textrm{-}1.0000 \ldots    1.0000$ &  $  \textrm{-}1.0000 \ldots    1.0000$ \\
$s^{(3)}$  &                                        &  $  \textrm{-}1.0000 \ldots    1.0000$ &  $  \textrm{-}0.4799 \ldots    0.4795$ &  $  \textrm{-}5.0000 \ldots    3.4373$ &  $  \textrm{-}3.5129 \ldots    1.8573$ &  $  \textrm{-}1.8748 \ldots    2.0801$ \\
$s^{(4)}$  &                                        &  $ \textrm{-}21.9001 \ldots   21.2146$ &  $  \textrm{-}1.0000 \ldots    1.0000$ &  $  \textrm{-}0.0000 \ldots    0.0000$ &  $  \textrm{-}0.0000 \ldots    0.0000$ &  $  \textrm{-}0.0000 \ldots    0.0000$ \\
$s^{(5)}$  &                                        &                                        &  $  \textrm{-}1.0000 \ldots    1.0000$ &  $  \textrm{-}5.0000 \ldots    1.6368$ &  $  \textrm{-}3.6400 \ldots    0.7000$ &  $  \textrm{-}1.9558 \ldots    1.3548$ \\
$s^{(6)}$  &                                        &                                        &  $ \textrm{-}10.0000 \ldots   10.0000$ &  $  \textrm{-}3.1416 \ldots    3.1416$ &  $  \textrm{-}3.1416 \ldots    0.0000$ &  $  \textrm{-}3.1416 \ldots    0.0000$ \\
$s^{(7)}$  &                                        &                                        &  $  \textrm{-}1.2745 \ldots    1.2701$ &  $  \textrm{-}1.5708 \ldots    1.5342$ &  $  \textrm{-}1.5708 \ldots    1.0625$ &  $  \textrm{-}1.5708 \ldots    1.0959$ \\
$s^{(8)}$  &                                        &                                        &  $ \textrm{-}10.0000 \ldots   10.0000$ &  $  \textrm{-}1.3921 \ldots    2.1682$ &  $  \textrm{-}2.2274 \ldots    1.5482$ &  $  \textrm{-}5.0000 \ldots    1.1894$ \\
$s^{(9)}$  &                                        &                                        &                                        &  $  \textrm{-}5.0000 \ldots    5.0000$ &  $  \textrm{-}5.0000 \ldots    4.6218$ &  $  \textrm{-}3.6561 \ldots    3.8614$ \\
$s^{(10)}$ &                                        &                                        &                                        &  $  \textrm{-}1.3917 \ldots    1.8215$ &  $  \textrm{-}1.5703 \ldots    1.8112$ &  $  \textrm{-}4.8688 \ldots    2.1159$ \\
$s^{(11)}$ &                                        &                                        &                                        &  $  \textrm{-}5.0000 \ldots    5.0000$ &  $  \textrm{-}4.2228 \ldots    4.3884$ &  $  \textrm{-}5.0000 \ldots    3.7174$ \\
$s^{(12)}$ &                                        &                                        &                                        &  $  \textrm{-}3.2458 \ldots    2.1586$ &  $  \textrm{-}3.7981 \ldots    1.5704$ &  $  \textrm{-}3.7094 \ldots    4.0843$ \\
$s^{(13)}$ &                                        &                                        &                                        &  $  \textrm{-}5.0000 \ldots    5.0000$ &  $  \textrm{-}4.9543 \ldots    3.1231$ &  $  \textrm{-}5.0000 \ldots    5.0000$ \\
$s^{(14)}$ &                                        &                                        &                                        &  $  \textrm{-}0.0000 \ldots    1.0000$ &  $  \textrm{-}2.8370 \ldots    1.2062$ &  $  \textrm{-}2.7263 \ldots    1.6122$ \\
$s^{(15)}$ &                                        &                                        &                                        &                                        &  $  \textrm{-}4.9879 \ldots    4.1318$ &  $  \textrm{-}5.0000 \ldots    4.6168$ \\
$s^{(16)}$ &                                        &                                        &                                        &                                        &  $  \textrm{-}1.7225 \ldots    1.7902$ &  $  \textrm{-}5.0000 \ldots    3.2957$ \\
$s^{(17)}$ &                                        &                                        &                                        &                                        &  $  \textrm{-}3.7233 \ldots    4.2734$ &  $  \textrm{-}5.0000 \ldots    5.0000$ \\
$s^{(18)}$ &                                        &                                        &                                        &                                        &  $  \textrm{-}4.0686 \ldots    1.5315$ &  $  \textrm{-}3.9515 \ldots    3.3214$ \\
$s^{(19)}$ &                                        &                                        &                                        &                                        &  $  \textrm{-}5.0000 \ldots    2.7369$ &  $  \textrm{-}5.0000 \ldots    5.0000$ \\
$s^{(20)}$ &                                        &                                        &                                        &                                        &  $  \textrm{-}0.0000 \ldots    1.0000$ &  $  \textrm{-}0.0000 \ldots    1.0000$ \\
$s^{(21)}$ &                                        &                                        &                                        &                                        &  $  \textrm{-}0.0000 \ldots    1.0000$ &  $  \textrm{-}0.0000 \ldots    1.0000$ \\
$s^{(22)}$ &                                        &                                        &                                        &                                        &                                        &  $  \textrm{-}0.0000 \ldots    1.0000$ \\
$s^{(23)}$ &                                        &                                        &                                        &                                        &                                        &  $  \textrm{-}0.0000 \ldots    1.0000$ \\
$s^{(24)}$ &                                        &                                        &                                        &                                        &                                        &  $  \textrm{-}0.0000 \ldots    1.0000$ \\
$s^{(25)}$ &                                        &                                        &                                        &                                        &                                        &  $  \textrm{-}0.0000 \ldots    1.0000$ \\
\bottomrule
\end{tabular}
   \end{adjustbox}
  \caption{The calculation of $\xurel$ requires defined state space
    limits for each environments. However, some environments define
    the limits as $(-\infty,\infty)$.  In these cases we collected
    state space samples and defined the limits empirically.  }
\end{table}

\end{document}